\pdfoutput=1

\documentclass[11pt]{article}

\usepackage[preprint]{acl}

\usepackage{times}
\usepackage{latexsym}

\usepackage[T1]{fontenc}
\usepackage{placeins}

\usepackage[utf8]{inputenc}

\usepackage{microtype}

\usepackage{inconsolata}

\usepackage{graphicx}

\usepackage{booktabs}
\usepackage{subcaption}
\usepackage{xcolor}         
\usepackage{amsmath}

\definecolor{darkblue}{rgb}{0, 0, 0.5}
\hypersetup{colorlinks=true, citecolor=darkblue, linkcolor=darkblue, urlcolor=darkblue}
\usepackage{graphicx} 
\usepackage{cleveref}

\title{Chained Tuning Leads to Biased Forgetting}

\author{
 \textbf{Megan Ung}, \enspace
 \textbf{Alicia Sun}, \enspace
 \textbf{Samuel J. Bell}, \enspace
 \textbf{Bhaktipriya Radharapu}, \enspace
\\
 \textbf{Levent Sagun}, \enspace
 \textbf{Adina Williams}
\\
 Meta FAIR,
\\
 \texttt{\small \{meganu, aliciasun, sjbell, bhakti, leventsagun, adinawilliams\}@meta.com} \\
}

\begin{document}
\maketitle
\begin{abstract}
Large language models (LLMs) are often fine-tuned for use on downstream tasks, though this can degrade capabilities learned during previous training. 
This phenomenon, often referred to as catastrophic forgetting, has important potential implications for the safety of deployed models. 
In this work, we first show that models trained on downstream tasks forget their safety tuning to a greater extent than models trained in the opposite order.
Second, we show that forgetting disproportionately impacts safety information about certain groups. 
To quantify this phenomenon, we define a new metric we term \emph{biased forgetting}. We conduct a systematic evaluation of the effects of task ordering on forgetting and apply mitigations that can help the model recover from the forgetting observed. 
We hope our findings can better inform methods for chaining the finetuning of LLMs in continual learning settings to enable training of safer and less toxic models. 
\end{abstract}

\section{Introduction}

\textit{Catastrophic forgetting}---the loss of information gained in earlier rounds of training as a consequence of subsequent rounds of training \citep{mccloskey1989catastrophic, ratcliff1990connectionist}---can pose a challenge in the context of ML model development \citep{goodfellow2014,kirkpatrick-etal-2016, kemker-etal-2018}. Recent works have also found evidence of catastrophic forgetting in the context of large language models (LLMs) \citep{Kotha2023UnderstandingCF, Luo2023AnES, razdaibiedina-etal-2023, Li2024ExaminingFI}. 
While finetuning with methods such as reinforcement learning from human feedback and instruction-tuning have been shown to be helpful for guiding models towards generating more desirable outputs \citep{Bai2022TrainingAH}, LLMs can still be brittle when finetuned on subsequent tasks.
For example, previous work has shown that adversarial testing or red teaming can bypass safety mechanisms \citep{Perez2022RedTL}, and safety metrics can degrade even when the model is subsequently fine tuned on benign downstream tasks \citep{Qi2023FinetuningAL}. 

Despite empirical observations of catastrophic forgetting, it remains unclear which post-training recipes can lead to forgetting. In this paper, we study how different finetuning regimes, including task ordering and learning rate, influence the severity of catastrophic forgetting. Specifically, we investigate a new sub-phenomenon of catastrophic forgetting that we call \textbf{biased forgetting}, whereby model performance degrades disproportionately for safety-related tasks, or for specific demographic groups, and propose a simple metric to quantify it. Empirically, we found that the order of finetuning on different tasks matters: safety/bias task are more prone to forgetting when subsequently finetuned with a capability task as compared to the other order.
We hypothesized that certain tasks may be more prone to forgetting due to the width of the minima obtained the end of training, and empirically observed that safety tasks converged to sharper minima. Based on these observations, we explored two approaches for mitigating biased forgetting: 
determining task ordering using first task loss curvature, and retraining using a small portion of the forgotten task.

We hope these findings provide empirical guidelines for optimizing the finetuning process of LLMs to mitigate biased forgetting, and encourage more careful investigation into the causes of biased forgetting in LLM training. 


\section{Background}\label{sec:background}
\begin{figure}[ht]
  \centering
  \includegraphics[width=\linewidth]{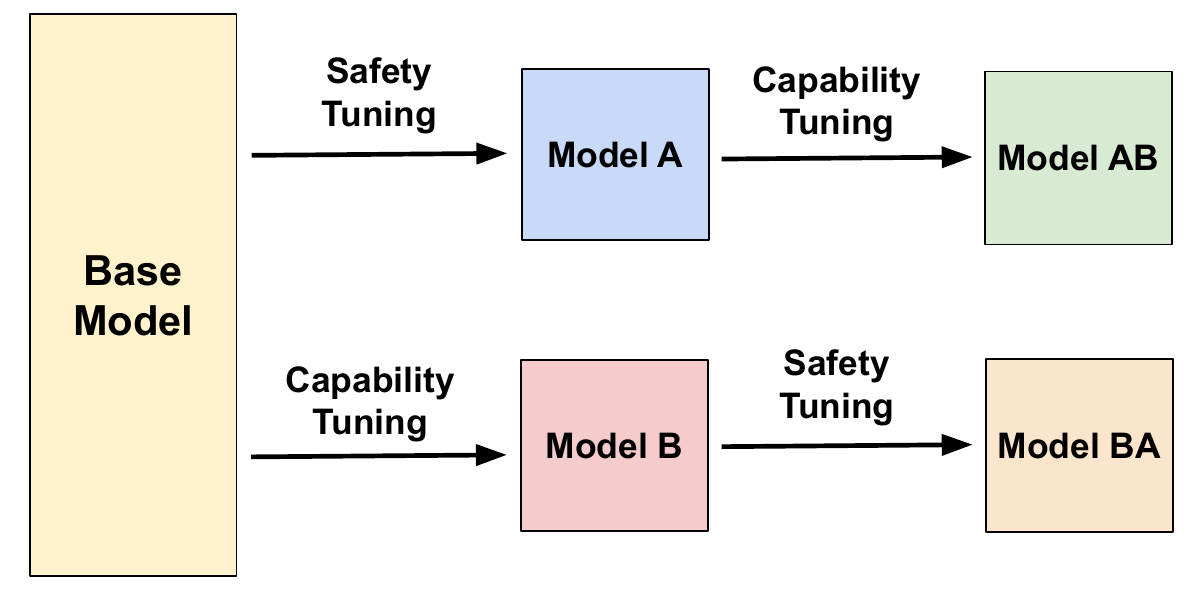}
  \caption{Task ordering experimental set up. Finetuning on a safety task first and then a capability task, and vice versa.}
  \label{fig:diagram-task-ordering}
\end{figure}
Under current, field-standard practice, the finetuning of LLMs is ``chained'', i.e., LLMs generally undergo multiple subsequent phases of finetuning, taking place sequentially, in the manner of continual learning \citep{wang2024comprehensive}. We broadly categorize finetuning as being either \emph{capability tuning} or \emph{safety tuning} depending on its primary intended purpose. In capability tuning, the model is tuned to improve its performance on some pre-specified task, such as scientific reasoning or factual question answering. 
In safety tuning, the LLM is finetuned to generate responses that are deemed safer, less biased, less toxic, or generally less harmful.\footnote{\label{safe_footnote}While there can be reasonable debate over what is and is not considered safer, biased, toxic or harmful, our goal is to demonstrate that the order of finetuning tasks when training LLMs negatively affects the forgetting of bias and safety tasks more than for capability tasks.}

To measure the disparity of forgetting, we define a new metric called \textit{biased forgetting} which measures the difference between the average and group forgetting across demographics. We investigate the effect of biased forgetting in controlled settings, keeping all other things equal in order to isolate the contributions of each LLM training decision. A controlled setting will enable us to better understand which design decisions impact the amount of biased forgetting, and determine how best to target mitigations. In particular, we focus here on the order of finetuning, the methods for finetuning, and the effect of different learning rate and batch size on forgetting and biased forgetting. 

With respect to tuning order, we adhere to the following approach to training LLMs, which involves two stages: (1) the pretraining stage where the model is trained to encode general-purpose representations via self-supervised learning on a large unlabeled text corpus, (2) the finetuning stage(s) where the model is trained on one or more smaller scale datasets in sequence, often with supervision to make the model more aligned to downstream tasks (via supervised finetuning) or human preferences (via reinforcement learning approaches). 
While safety tuning may be performed in-house by the original LLM developer as part of a model release process \citep{touvron2023llama, claude3}, capability tuning may also be performed ``downstream'' by a third-party, accessing opensource weights and performing additional rounds of finetuning to adapt the model to their specific needs.
Because this is a general use case for many openly available LLMs, understanding the effects of such downstream capability tuning on the upstream safety tuning is a primary motivation for our work. In particular, we would like to determine how much the upstream safety tuning is affected by downstream capability tuning. To do so, we explore the ordering of tasks in a controlled setting, holding data size constant.


In summary, we analyze the consequences of two main LLM training decisions on biased forgetting in a chained tuning, continual learning setting.
\begin{itemize}
   \item Task ordering (\S\ref{subsec:task-order}): We explore how the sequence in which tasks are presented affects the retention of previously learned information. We experiment with two sequences as in \Cref{fig:diagram-task-ordering}: (1) Task A is the capability task and Task B is the safety task, (2) Task A is the safety task and Task B is the capability task.
    \item Initial task learning paradigm (\S\ref{subsec:initial-learning-paradigm}): We vary the learning rate of both the first task and the second task to better understand the consequences of training hyper-parameter decisions on (biased) forgetting.
\end{itemize}

\section{Chained Tuning}
\subsection{Experimental Setup}
In this work, we use the state-of-the-art pre-trained large language model, LLaMa-v2 \citep{touvron2023llama}, as it is openly available and will enable our results to be reproduced. We use the LLaMa-v2 7B pre-trained model as our base model, as opposed to a chat-optimized version, because query refusal could impede our ability to characterize the effect of chaining finetuning and other finetuning hyperparameter decisions. 

Following standard practice in supervised finetuning, we use AdamW optimizer \citep{Loshchilov2017DecoupledWD} with $\beta_1=0.9$ and $\beta_2=0.95$. We use a linear learning rate scheduler with a weight decay of 0.1 and without any warmup steps. We use a sequence length of 2048, and finetuning for 100 optimization steps. We repeat each experiment three times with different random seeds, and report the average accuracy. For sequential tuning, we pick the best model for the first task, and use it as the baseline model for the second task. Unless otherwise noted, we report the majority of our results using learning rate of $1e^{-5}$ and batch size of 16 for ordering and tuning methods. 

\subsection{Tasks and Datasets}
To isolate the effect of task template and style, we reformat most of our tasks into QA format and finetune the model using instruction tuning. This enables maximal comparability across capability and safety tasks. We append an instruction to the front of each prompt, where the instruction is shown in Appendix~\S\ref{subsec:sft-template}. We sub-sample all dataset to have the same number of examples (2261). We also use random stratified sampling to ensure same number of examples from each demographic group and each label class.

\paragraph{Capability Tasks.} We use three capability tasks: (1) AI2 Reasoning Challenge (ARC) dataset \citep{clark2018think}, (2) CommonsenseQA (CQA) dataset \citep{Talmor2019CommonsenseQAAQ}, and (3) CommonsenseQA 2.0 (CQA2) \citep{Talmor2021CommonsenseQA2E}. 

ARC contains grade-school science questions where each question has 4 choices. We use the ARC easy set for finetuning (referring to it as `ARC' from here on, unless otherwise specified). CQA is a dataset for commonsense question answering dataset, where each question has 5 choices. CQA2 is a recent challenging QA dataset collected
with a model-in-the-loop approach, and each question is binary with only yes/no choices. All of the capability tasks are finetuned and evaluated on restricted output space, where the model responds with "The correct answer is [answer\_letter]" given the question and choices. 

\paragraph{Safety and Bias Tasks.} While there are many existing safety datasets that could be used to explore the phenomenon of biased forgetting, we focus three safety tasks that are standard in the field and encode different operationalizations of safety: (1) ToxiGen \citep{hartvigsen2022toxigen}, (2) Bias Benchmark for QA (BBQ, \citealt{parrish-etal-2022-bbq}), and (3) SaFeRDialogues \citep{ung2022saferdialogues}.

The ToxiGen dataset contains both toxic and benign statements about 13 demographic groups. For ToxiGen, we use the revised dataset (``v2'') from \cite{hosseini2023empirical} that builds on the original ToxiGen dataset and reduces noise by filtering out sentences for which annotators disagree on the target demographic group. We recast ToxiGen into QA format, which we denote as ToxiGenQA, and the model is trained to respond with [This is toxic/not toxic] given a prompt. 

The Bias Benchmark for QA dataset (BBQ; \citealt{parrish-etal-2022-bbq}) contains expert-written questions meant to emphasize and identify model social biases against people belonging to specific demographic groups. 
The dataset has 9 demographic groups and each question comes with either an ambiguous context (correct answer unknown) or a disambiguated context, where the disambiguated context provides additional information that is necessary to answer the question. 
We construct the training dataset with examples from BBQ's human validation sub-dataset, and fill the rest with the provided templated examples. We ensure an even split across all the social groups, and balance between disambiguated and ambiguous questions within each group. 

SaFeRDialogues dataset \citep{ung2022saferdialogues} containing safety failures, feedback, and graceful responses. To investigate if task format affects forgetting, we format SaFeRDialogues as a generative task with unlimited output space. To determine whether model responses can be deemed `safe', we evaluate the model's outputs using the ToxiGen classifier \citep{hartvigsen2022toxigen} tuned on RoBERTa \citep{liu2019roberta} to score continuations given the safety failures prompts\footnote{see footnote 1}.

\paragraph{Evaluation Metrics.}
To quantify the amount of forgetting we observe, we propose the following metrics. These metrics will enable us to track the impact of forgetting overall, and elucidate disparate impact, if any, on particular demographic groups. Let $\theta^*_{A}$ be a model tuned on task $A$ only, and let $\theta^*_{AB}$ be a model tuned sequentially on task $A$ followed by task $B$.
Then, we define forgetting for task $A$ as difference in task $A$ performance after training on task B,
\begin{equation}
    \mathrm{Forgetting}_{AB} = \mathrm{Acc}_A(\theta^*_{A})-\mathrm{Acc}_A(\theta^*_{AB}) \ ,
\end{equation}
where $\mathrm{Acc}_i(\cdot)$ is task $i$ accuracy. 
As both BBQ and ToxiGen are equipped with group information, we can evaluate group-disaggregated performance throughout tuning.
We define the per-group forgetting as
\begin{equation*}
    \mathrm{Forgetting}_{A,g} = \mathrm{Acc}_A(\theta^*_{A},g)-\mathrm{Acc}_A(\theta^*_{AB},g) \ ,
\end{equation*}
where $\mathrm{Acc}_i(\cdot, g)$ is the accuracy for samples in group $g$ on task $i$. We additionally report worst group (WG) forgetting, which indicates the highest level of forgetting of a single group, $\max_{g} \{\mathrm{Forgetting}_{A,g}\} $.
In some cases, we calculate and report \textit{relative} forgetting where 
\begin{equation*}
    \mathrm{Relative Forgetting}_{AB} = \frac{\mathrm{Forgetting}_{AB} }{ \mathrm{Acc}_A(\theta^*_{A},g)}
\end{equation*}
Finally, we define \emph{biased forgetting} as the gap between the per-group forgetting for group $g$ on task $A$ and the overall forgetting on task $A$, as 
\begin{align}
\begin{aligned}
    \mathrm{BiasedForgetting}_{A,g} &=  \\ \mathrm{Forgetting}&_{A,g}-\mathrm{Forgetting}_A \ .
\end{aligned}
\end{align}
While we typically report the \emph{maximum} BiasedForgetting (i.e., the gap between the worst-group and the overall) in our experiments below, this metric is flexible and can be adapted to cover an arbitrary number of worst-groups as needed.

\section{Results}

\subsection{Task order matters}\label{subsec:task-order}

\begin{table*}[!ht]
\begin{tabular}{lccccccc}
\toprule
\textbf{Capability} & \textbf{Safety}  & \multicolumn{3}{c}{\textbf{Capability$\rightarrow$Safety}}  & \multicolumn{3}{c}{\textbf{Safety$\rightarrow$Capability}}     \\
                &             & $\mathrm{Acc}_{cap}$ & $\mathrm{Acc}_{safety}$ & $\mathrm{Forgetting}_{cap}$ & $\mathrm{Acc}_{cap}$ & $\mathrm{Acc}_{safety}$ & $\mathrm{Forgetting}_{safety}$ \\
                \midrule
ARC             & TQA         &    77.11          &     90.79            &     0.1                &     75.15         &     84.29            &       6.87                 \\
                & BBQ         &     76.19         &              97.65   &         1.02            &        77.64      &    54.56             &         43.45               \\
                & SD          &       76.83       &      98.10           &        0.38             &       75.95       &        94.25         &           3.21            \\
                \midrule
CQA             & TQA         &      70.32        &    90.38             &  -0.05                 &      72.81        &     87.18            &        3.98                \\
                & BBQ         &       53.4       &    97.75             &        16.87             &        71.32      &   51.16              &        46.85                \\
                & SD          &      70.3        &          96.7       &        -0.03             &       70.02        &        96.62        &           0.84             \\
                \midrule
CQA2            & TQA         &    51.79          &    91.04            &     9.29                &   60.60           &      53.11           &         38.05               \\
                & BBQ         &   53.71           &    98.18              &         7.37            &   59.43           &    84.49             &     13.52                   \\
                & SD          &       62.08       &          98.73       &         -1.00            &       60.17       &         93.10        & 4.36 \\
\bottomrule
\end{tabular}
\caption{Task performance (accuracy \%) and forgetting when tuning in different order (capability task$\rightarrow$safety task vs safety task$\rightarrow$capability task). Forgetting is measured with respect to the performance change on a task before and after fine-tuning on another task. In this experiment, we use the same training hyper-parameters for the first and second task (learning rate $1e^{-5}$, and batch size 16). }
\label{table:ordering_table}
\end{table*}

\begin{figure}[!ht]
    \centering
    \includegraphics[width=\linewidth]{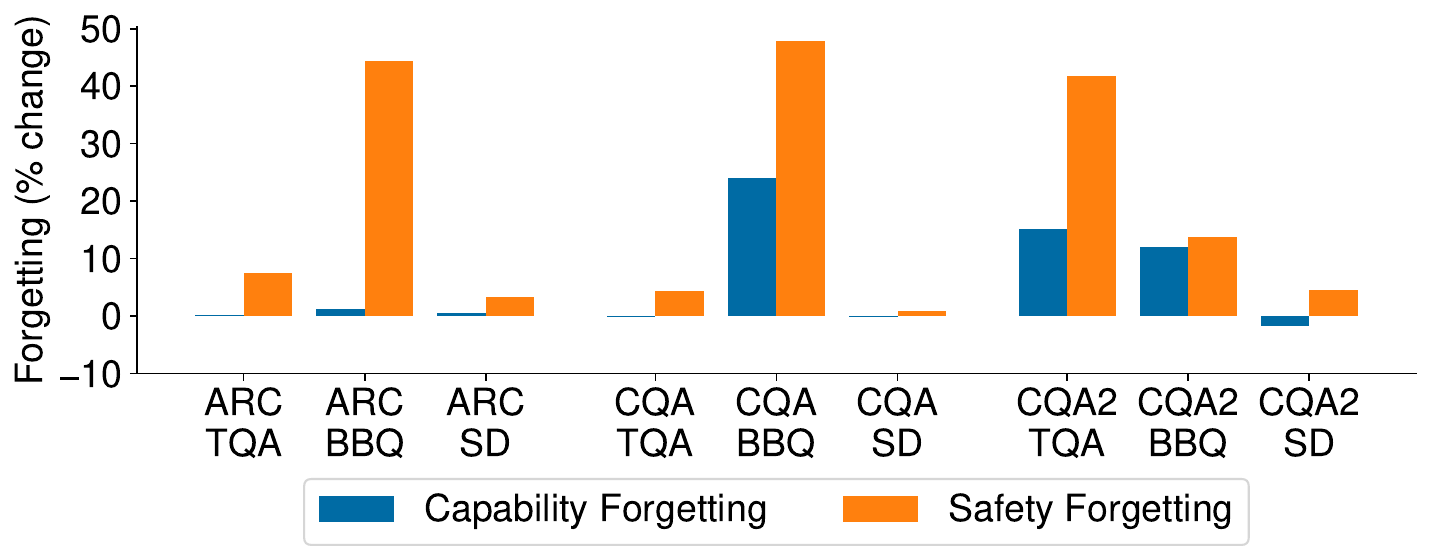}
    \caption{Relative Forgetting (\%) of the capability task and the safety task for each pair of tasks. Overall, there is more forgetting of the safety task (orange) than forgetting of the capability task (blue).}
    \label{fig:forgetting-safety-capability}
\end{figure}

We observe more forgetting on the first task when it is a safety task (ToxiGenQA, BBQ, SaFeRDialogues) compared to when it is capability task (ARC, CQA, CQA2) with supervised finetuning, as shown in \Cref{fig:forgetting-safety-capability}. We observe that this finding is consistent across different learning rates, as shown in Appendix-\Cref{fig:ordering_forgettingxlr}.
Interestingly, finetuning on safety tasks \emph{after} finetuning on capability tasks does not necessarily lead to catastrophic forgetting on the capability tasks (we even observe some negative $\mathrm{Forgetting}_{cap}$ in  \Cref{table:ordering_table}). 
In our case, capability task performance is largely maintained despite later safety tuning. 
We hypothesize that this could be due to task similarity. Some pairs of tasks are both in QA format, allowing the model to maintain the consistent response style learned during the first task. Additionally, we observe that SaferDialogues suffers from less forgetting compared to the other safety tasks in \Cref{table:ordering_table}. We believe this is due to several differences, as the task itself is a generation task and not in QA format like the other tasks used and the evaluation method for this task is dependent on the performance of the classifier used. 
This finding underscores the importance of task ordering and format similarity in sequential finetuning. 
\subsection{Forgetting is unevenly distributed}\label{subsec:biasedforgetting}

\begin{figure*}[!ht]
\centering
\begin{subfigure}{0.33\textwidth}
  \centering
  \includegraphics[width=\textwidth]{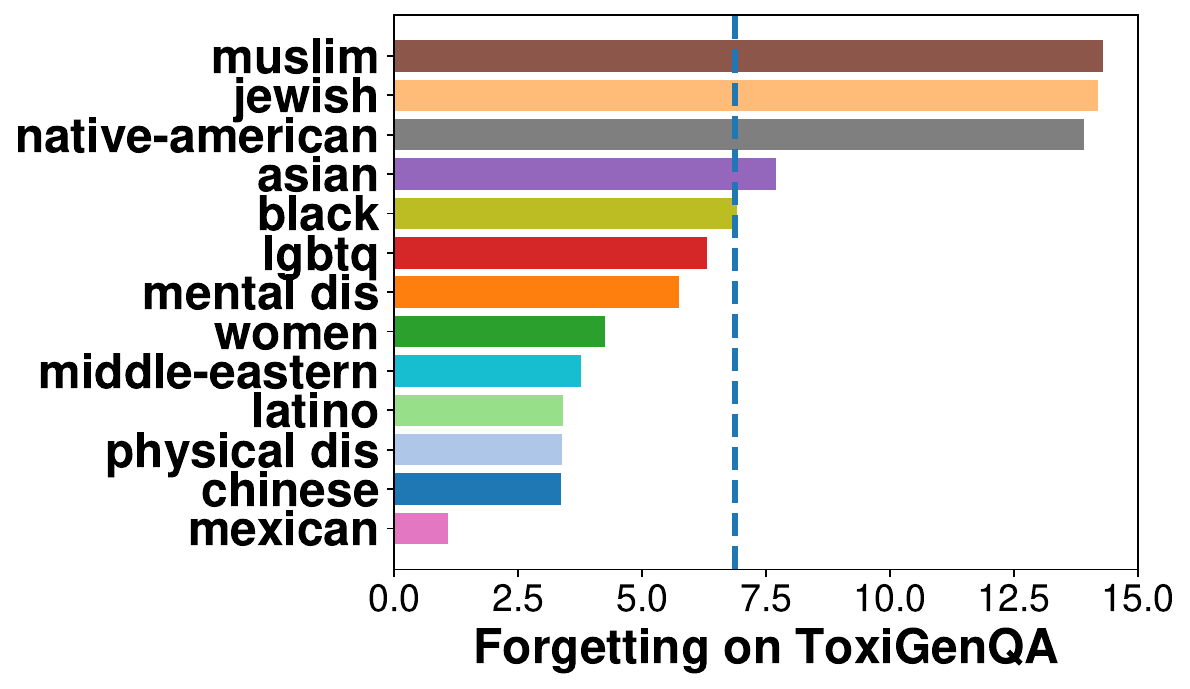}
    \caption{\label{fig:groupbyforget_sfttqasftarc}\centering TQA $\rightarrow$ ARC $\text{BiasedForgetting}_{max} = 7.42$}
\end{subfigure}\hfill
\begin{subfigure}{0.33\textwidth}
  \centering
  \includegraphics[width=\textwidth]{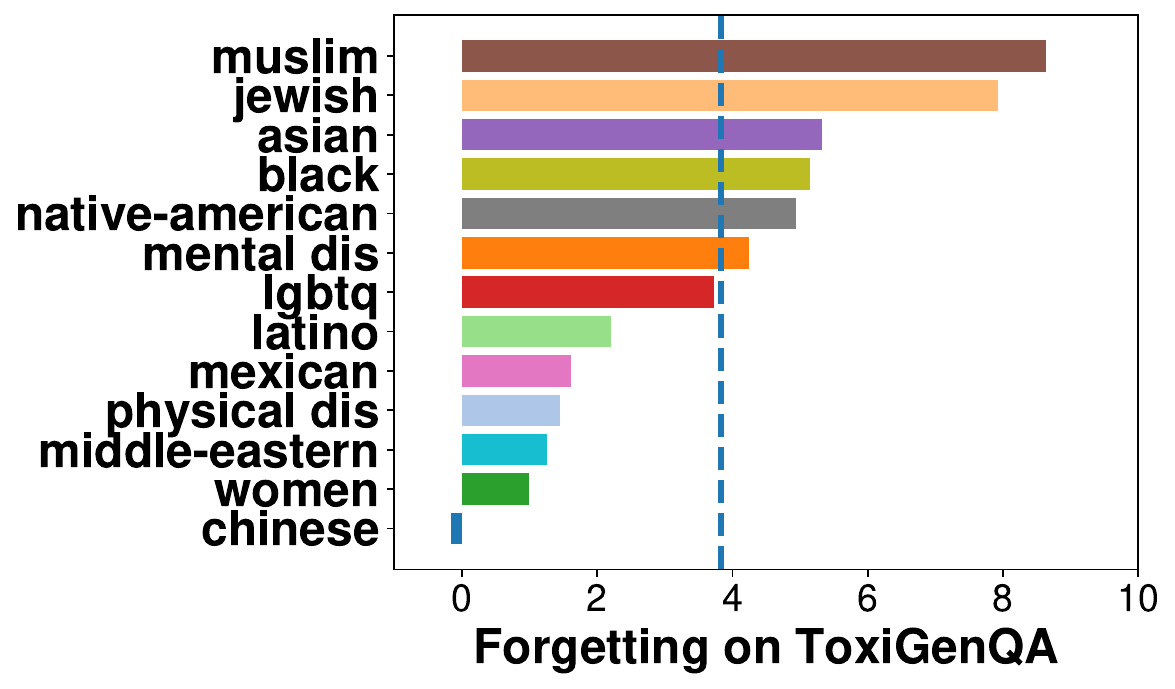}
    \caption{\label{fig:groupbyforget_sfttqasftcqa}\centering TQA $\rightarrow$ CQA $\text{BiasedForgetting}_{max} = 4.81$}
\end{subfigure}
\begin{subfigure}{0.33\textwidth}
  \centering
  \includegraphics[width=\textwidth]{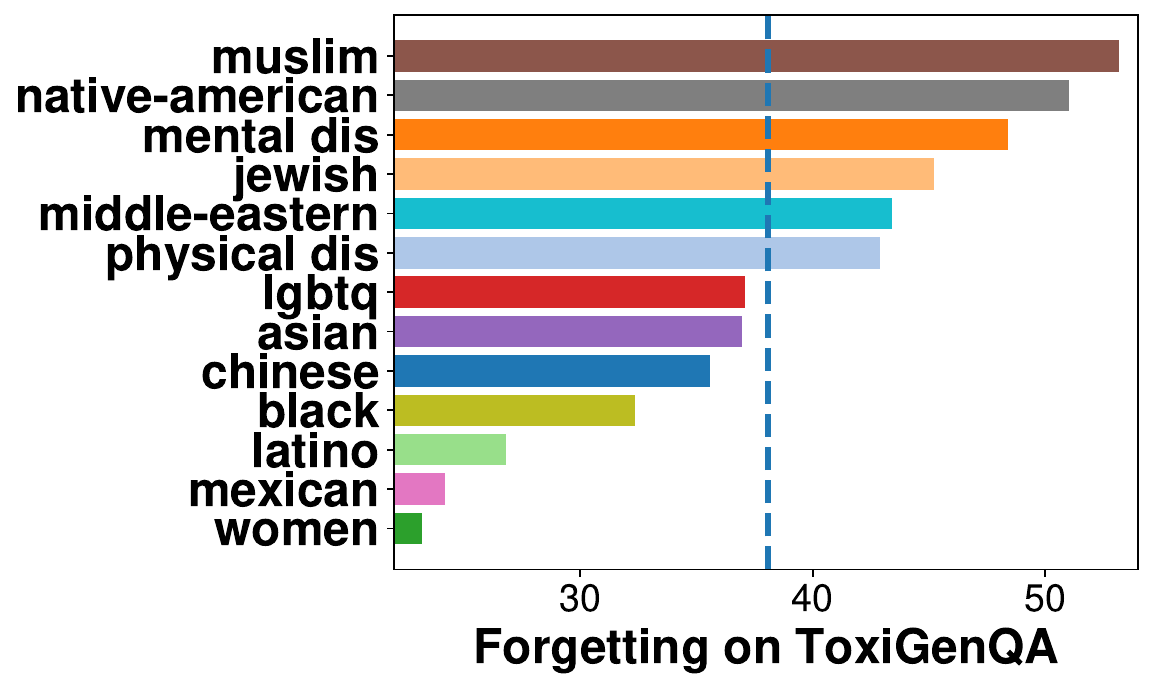}
  
    \caption{\label{fig:groupbyforget_sfttqasftcqa2}\centering TQA $\rightarrow$ CQA2 $\text{BiasedForgetting}_{max} = 10.09$}
\end{subfigure}
\begin{subfigure}{0.33\textwidth}
  \centering
  \includegraphics[width=\textwidth]{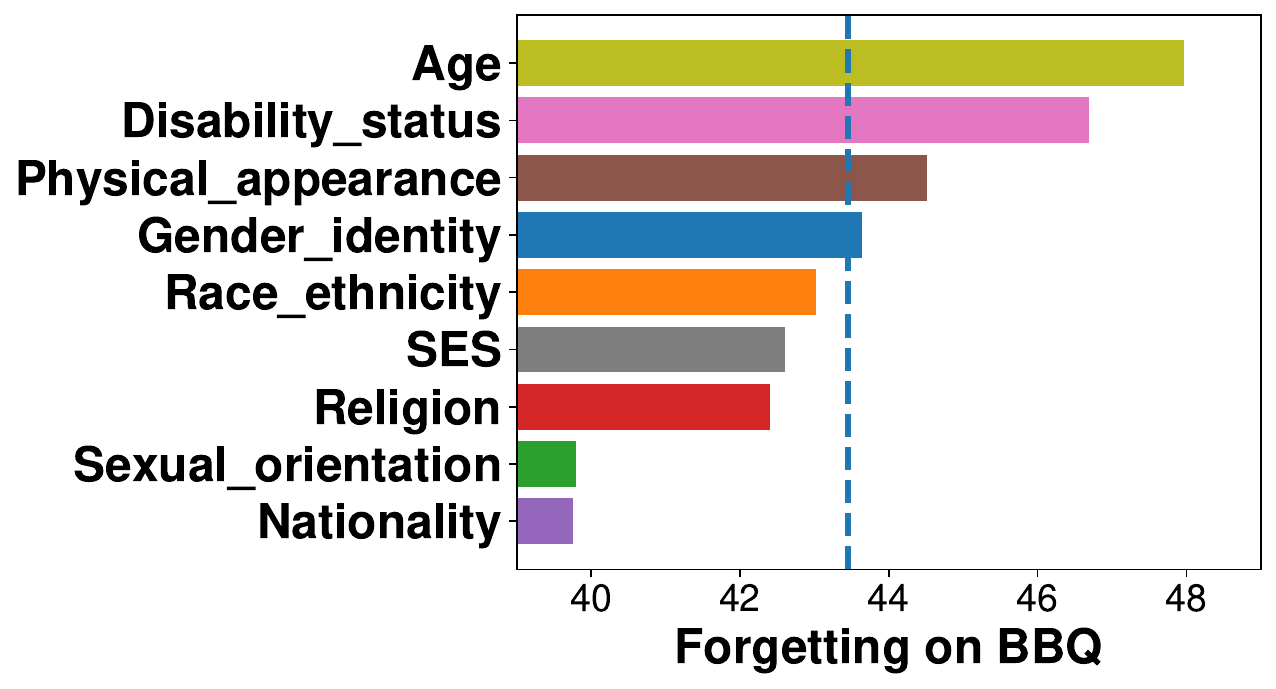}
    \caption{\label{fig:groupbyforget_sftbbqsftarc}\centering BBQ $\rightarrow$ ARC $\text{BiasedForgetting}_{max} = 4.52$}
\end{subfigure}\hfill
\begin{subfigure}{0.33\textwidth}
  \centering
  \includegraphics[width=\textwidth]{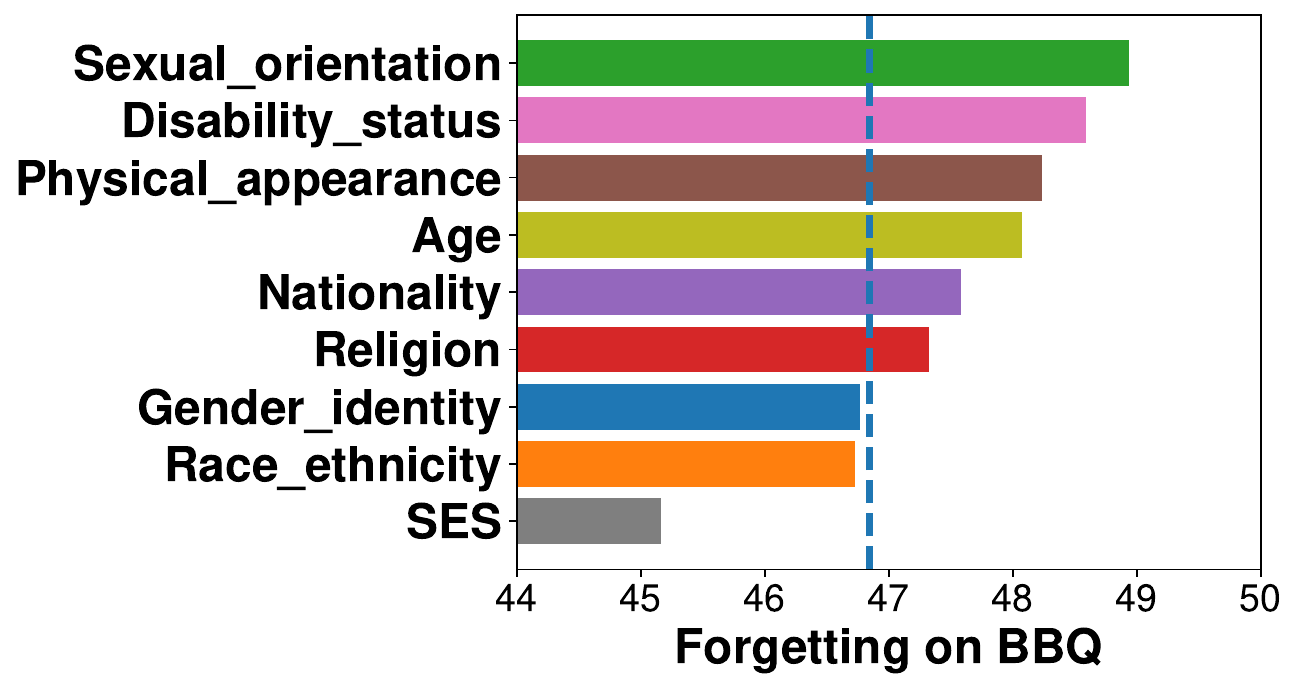}
    
    \caption{\label{fig:groupbyforget_sftbbqsftcqa}\centering BBQ $\rightarrow$ CQA $\text{BiasedForgetting}_{max} = 2.10$}
\end{subfigure}
\begin{subfigure}{0.33\textwidth}
  \centering
  \includegraphics[width=\textwidth]{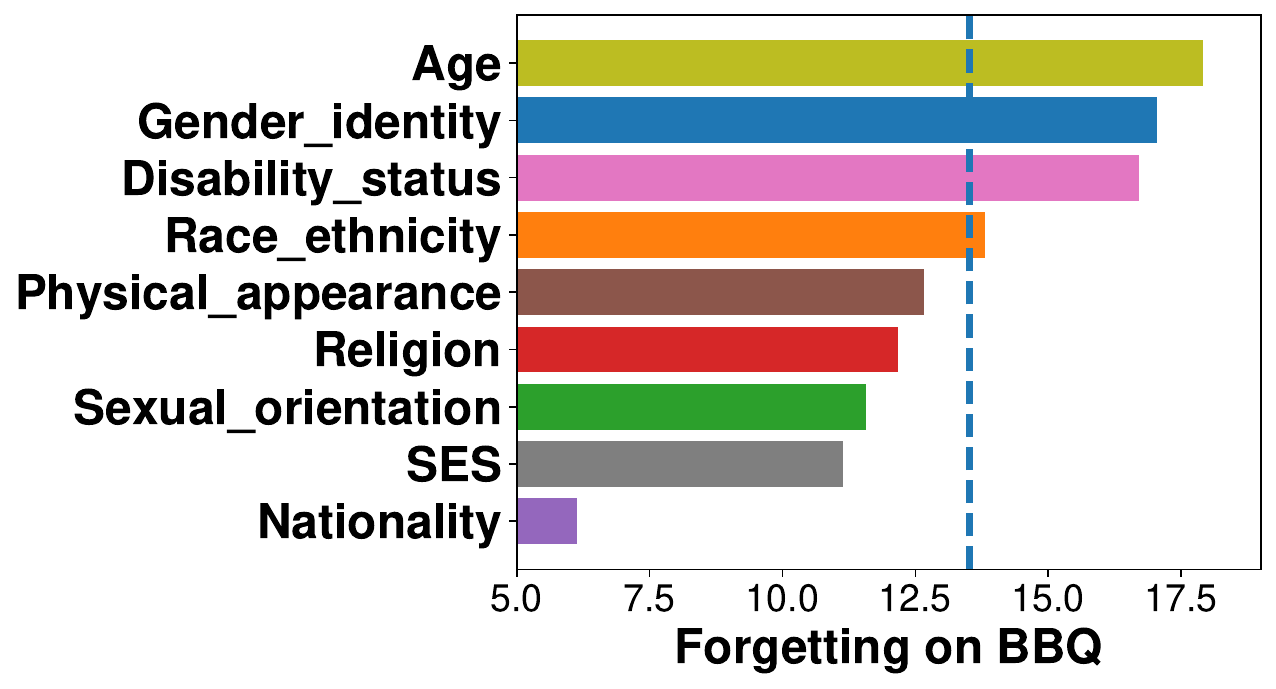}
    \caption{\label{fig:groupbyforget_sftbbqsftcqa2}\centering BBQ $\rightarrow$ CQA2 $\text{BiasedForgetting}_{max} = 4.39$}
\end{subfigure}
\caption{Forgetting by groups in ToxiGenQA (a, b, c) and BBQ (d, e, f) followed by finetuning on a capability task (ARC, CQA, CQA2). All experiments use learning rate $1e-5$ and batch size 16. The blue dotted vertical line denotes the average forgetting on the safety task (TQA or BBQ).}
\label{fig:group_forgetting}
\end{figure*}

We observe varying degrees of biased forgetting when comparing demographic groups in safety tasks after finetuning on a safety task (ToxiGenQA and BBQ) and then finetuning on a capability task (ARC, CQA, and CQA2) in \Cref{fig:group_forgetting}. All pairs of chained tuning tasks suffer from Biased Forgetting, with the largest observed Biased Forgetting at $10.09$ when the model is finetuned on ToxiGenQA and then finetuned on CommonSenseQA 2. We plot the corresponding biased forgetting by group results for ToxiGenQA in \Cref{fig:bias_forgetting_tqa} and BBQ in \Cref{fig:bias_forgetting_bbq}. In \Cref{fig:groupbyforget_sfttqasftarc}, \Cref{fig:groupbyforget_sfttqasftcqa}, and \Cref{fig:groupbyforget_sfttqasftcqa2}, we see that `Muslim' is the top group that suffers from the most forgetting after the model is finetuned on a capability task. We also observe that along with the `Muslim' group, `Jewish' and `Native American' groups are also present in the top 5 (out of 13) groups that suffer from the most forgetting after finetuning on various capability tasks. We use Kendall’s tau-b correlation coefficient, which is a non-parametric test based on ranks, to assess the relationship between the biased forgetting of groups after finetuning on different capability tasks. We find a reasonably consistent ranking of groups between TQA $\rightarrow$ ARC and TQA $\rightarrow$ CQA (Kendall's tau $= 0.69$) and between TQA $\rightarrow$ ARC and TQA $\rightarrow$ CQA2 (Kendall's tau $= 0.46$). For the 3 task pairs where we finetune on BBQ and then a capability task (\Cref{fig:groupbyforget_sftbbqsftarc}, \Cref{fig:groupbyforget_sftbbqsftcqa}, and \Cref{fig:groupbyforget_sftbbqsftcqa2}), we see some more variation in the ordering though also notice that `Age' and `Disability Status' groups stay in the top 4 (of 9), and SES and `Religion' groups stay in the bottom 4. We also find a reasonably consistent ranking of groups between BBQ $\rightarrow$ ARC and BBQ $\rightarrow$ CQA2 (Kendall's tau $= 0.72$). We notice slightly more variation of ordering of the forgetting by groups in BBQ than in ToxiGenQA task pairs which may be due to the reasoning ability required of the BBQ task which we discuss more in Appendix~\S\ref{subsec:dataset-bbq-discussion}. 

\subsection{Wide minima are less forgotten}

\label{subsec:curvature}

Why are certain tasks forgotten more?  Previous work has suggested a relationship between the curvature of the minima obtained at the end of training and both generalization performance \citep{Hochreiter1997, jastrzebski2018three} and catastrophic forgetting \citep{Mirzadeh2020UnderstandingTR}.
Here, we investigate whether first task minima curvature can explain the effect of task ordering observed in \cref{subsec:task-order}.

Given a model $\theta^{*}_{AB}$ trained sequentially on $A$ followed by $B$, we follow \citet{Mirzadeh2020UnderstandingTR} in using a Taylor expansion to approximate the change in first task loss $L_A$,
\begin{equation}
    \label{eq:taylor}
    L_A(\theta^*_{AB}) - L_A(\theta^*_A) \approx \frac{1}{2} \Delta \mathbf{\theta}^{*\top}H\Delta\mathbf{\theta}^{*} ,
\end{equation}
where $H = \nabla^{2}L_A(\theta^*_A)$ is the Hessian of the first task loss.
\Cref{eq:taylor} relies on the assumption that the first task model $\theta^*_A$ is at, or reasonably near to, a local minima, which we validate by inspecting both training losses and the norms of gradient updates.
We can then bound the change in first task loss using the magnitude of the parameter change and the curvature of the minimum,
\begin{equation}
    \frac{1}{2} \Delta \mathbf{\theta}^{*\top}H\Delta\mathbf{\theta}^{*} \leq \frac{1}{2} \rho(H) ||\Delta\theta^*||^2 ,
\label{eq:loss_bound}
\end{equation}
where $\rho(H)$ is the spectral radius, or dominant eigenvalue, of the Hessian. 
Thus we see that two factors contribute to forgetting: the curvature after the first task, and the magnitude of parameter change during the second task.

We use power iteration to numerically approximate the dominant eigenvalue, making use of the Hessian-vector product trick to avoid computing the intractably large Hessian, using a modified version of a library by \cite{hessian-eigenthings}.
To improve efficiency we compute the Hessian using 50 random samples from the training set.
We compute the spectral radius after tuning on the first task only, for all safety and capability tasks.

\begin{figure*}[!ht]
    \centering
    \begin{subfigure}{0.48\textwidth}
        \centering
        \includegraphics[width=0.8\textwidth]{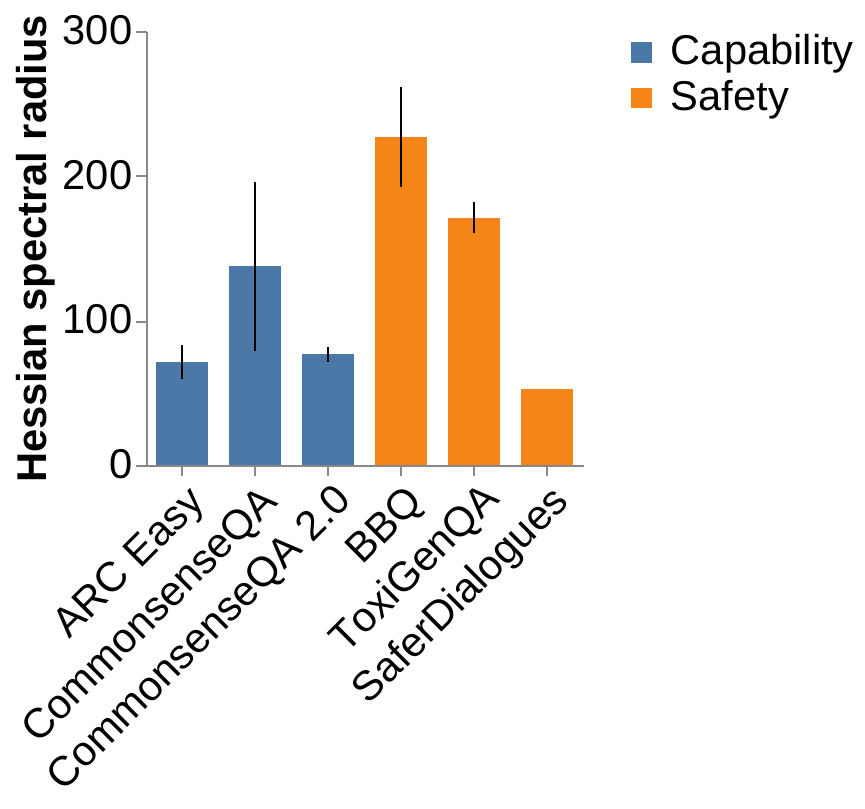}
        \label{fig:curvature-ordering}
    \end{subfigure}
    \hfill
    \begin{subfigure}{0.48\textwidth}
        \centering
        \includegraphics[width=\textwidth]{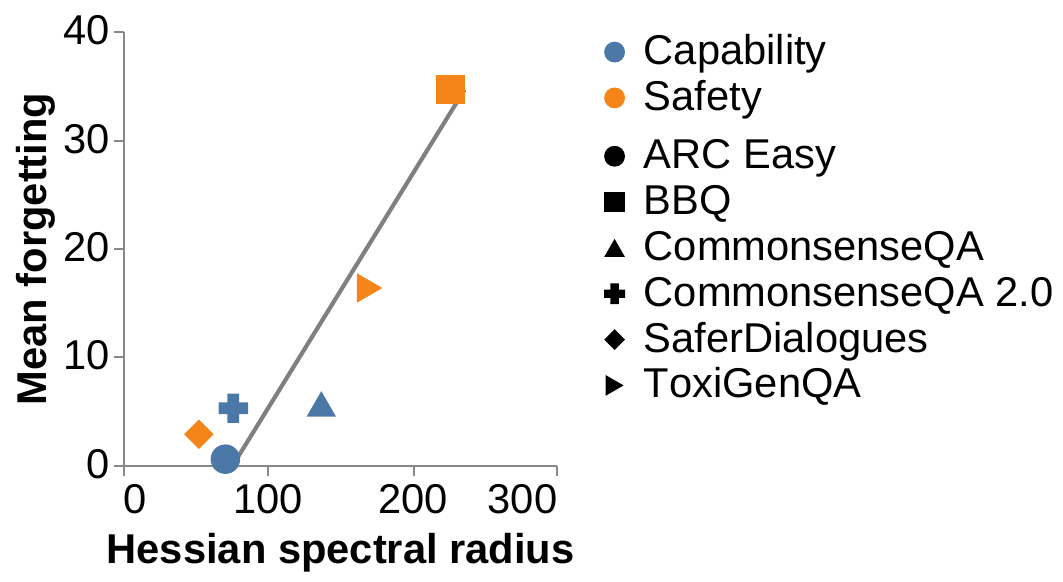}
        \label{fig:curvature-by-forgetting}
        \vspace{30pt}
    \end{subfigure}
    \hfill
    \caption{\textbf{(a)} Minima curvature (i.e. approx.\ spectral radius of the Hessian) obtained after training on the first task. \textbf{(b)} Mean downstream forgetting as a function of first task curvature. \textbf{Curvature explains a significant proportion of downstream forgetting.} Error bars are standard deviation over three training runs.}
    \label{fig:curvature}
\end{figure*}

\Cref{fig:curvature-ordering} shows a markedly sharper minima (i.e., larger spectral radii) for two of the three safety tasks, BBQ and ToxiGenQA, when compared with the minima obtained during capability training.
This could explain why training on safety tasks first may produce greater forgetting than when training on capability tasks first. 
The lower curvature for SaferDialogues is in line with the results in \cref{table:ordering_table} suggesting that training in SaferDialogues leads to less forgetting than other safety tasks. 
In \Cref{fig:curvature-by-forgetting}, we see that tasks leading to a wide minima result in significantly reduced downstream forgetting (OLS; $p\leq0.01$; $R^2=0.85$).
This might suggest that employing methods to bias training towards a wider minima may be beneficial in reducing safety forgetting. 


\subsection{Effect of learning rate}\label{subsec:initial-learning-paradigm}

\begin{figure*}[!ht]
\centering
\begin{subfigure}{0.3\textwidth}
  \centering
  \includegraphics[width=\textwidth]{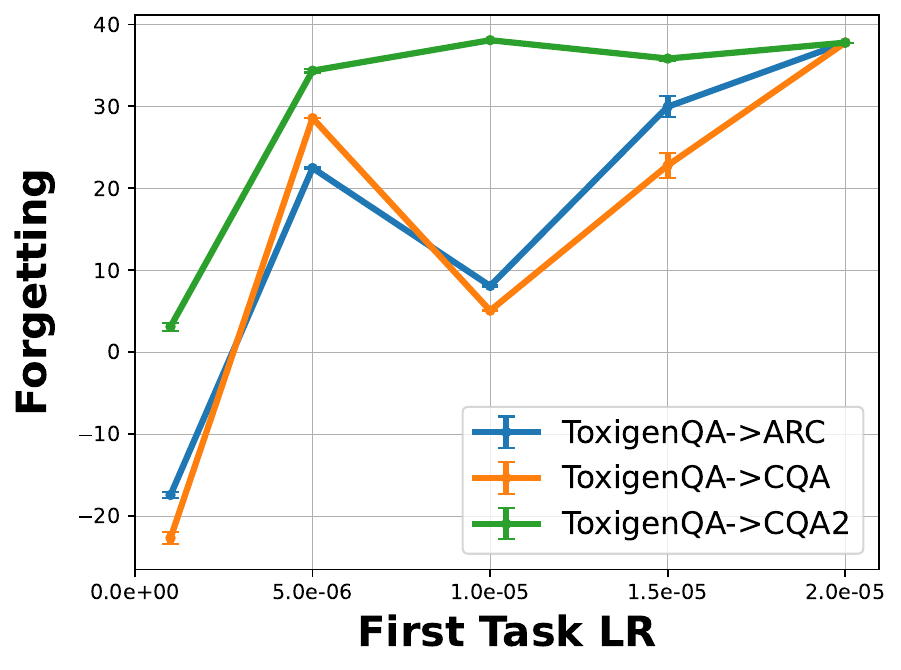}
    \label{fig:first_lr_forgetting_toxigen}
\end{subfigure}
\begin{subfigure}{0.3\textwidth}
  \centering
  \includegraphics[width=\textwidth]{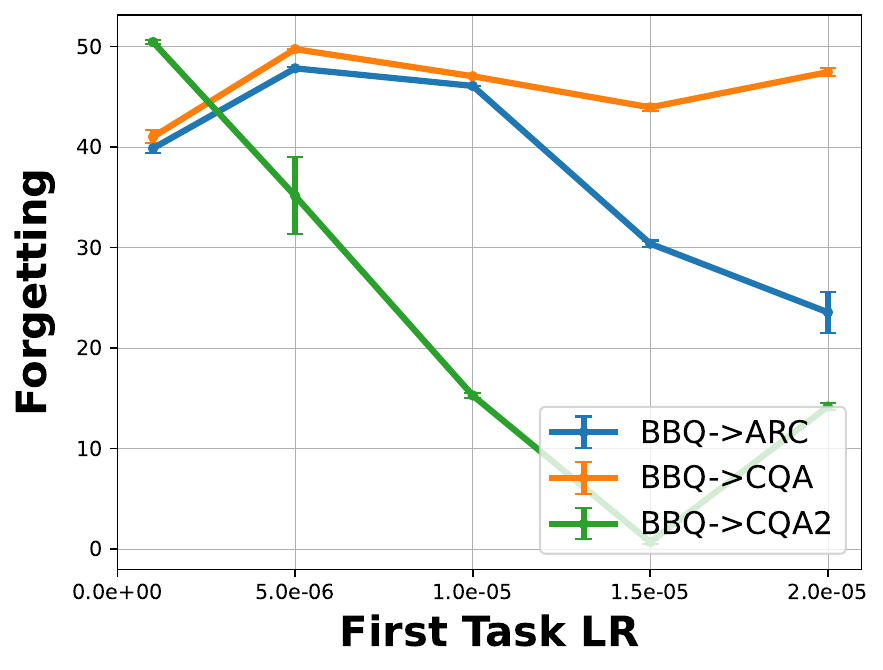}
    \label{fig:first_lr_forgetting_bbq}
\end{subfigure}
\begin{subfigure}{0.3\textwidth}
  \centering
  \includegraphics[width=\textwidth]{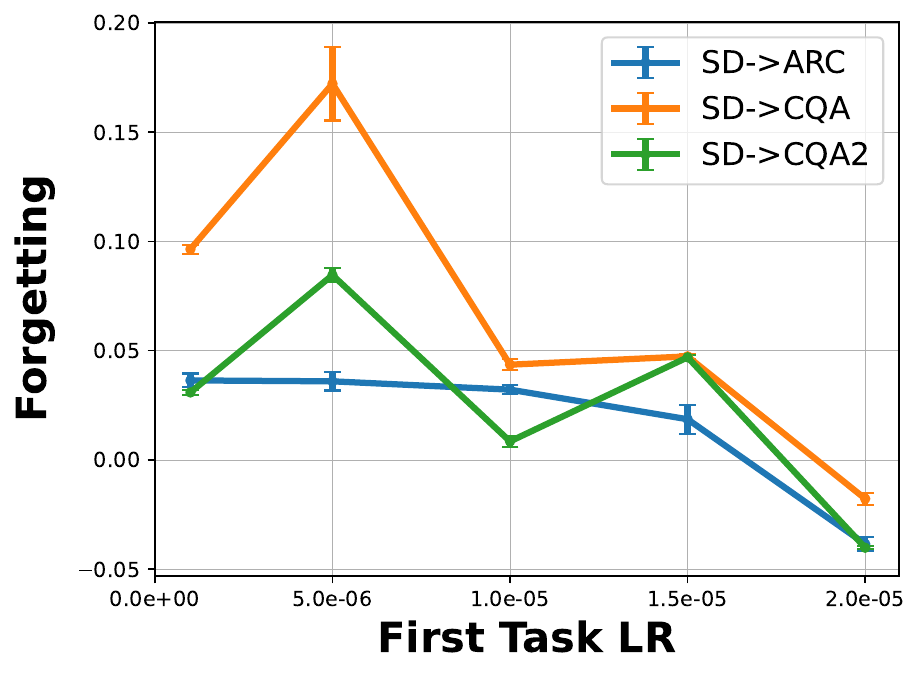}
    \label{fig:first_lr_forgetting_sft}
\end{subfigure}
\caption{Forgetting when varying first task learning rate for different safety task$\rightarrow$capability task sequence.}
\label{fig:first_lr_forgetting}
\end{figure*}

\begin{figure*}[!ht]
\centering
\begin{subfigure}{0.3\textwidth}
  \centering
  \includegraphics[width=\textwidth]{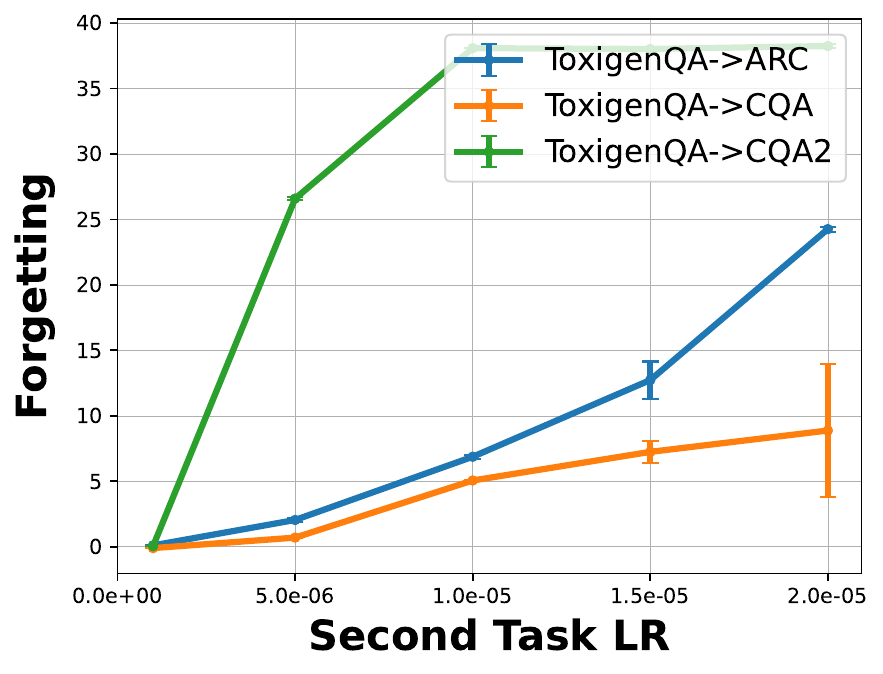}
    \label{fig:second_lr_forgetting_toxigen}
\end{subfigure}
\begin{subfigure}{0.3\textwidth}
  \centering
  \includegraphics[width=\textwidth]{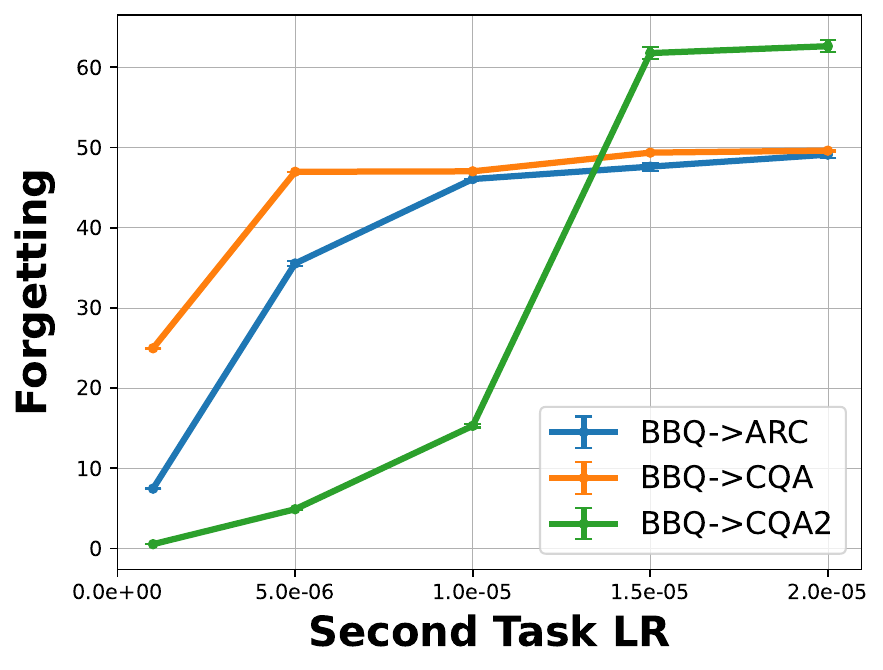}
    \label{fig:second_lr_forgetting_bbq}
\end{subfigure}
\begin{subfigure}{0.3\textwidth}
  \centering
  \includegraphics[width=\textwidth]{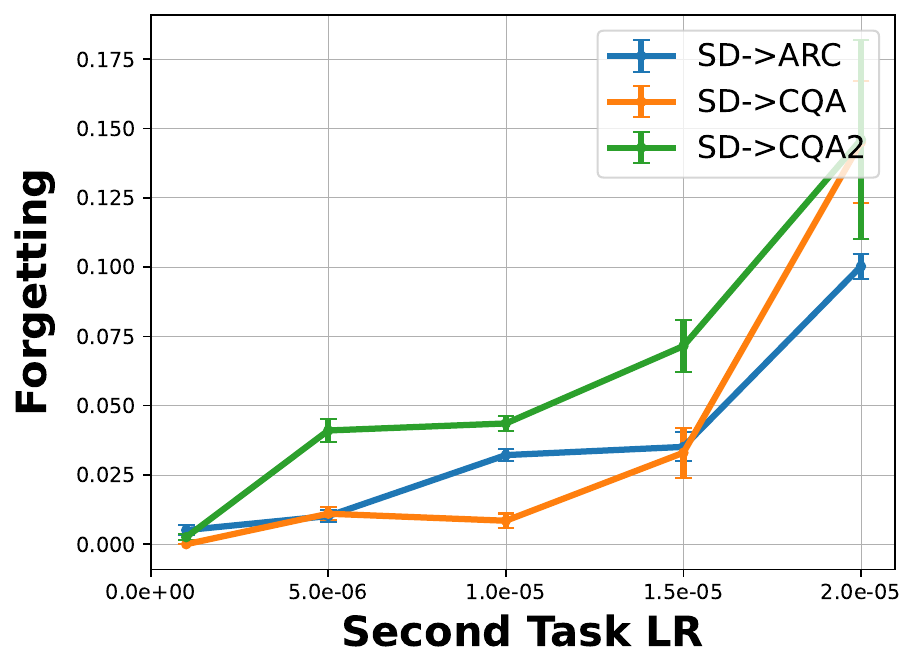}
    \label{fig:second_lr_forgetting_sd}
\end{subfigure}
\caption{Forgetting when varying second task learning rate for different safety task$\rightarrow$capability task sequence.}
\label{fig:second_lr_forgetting}
\end{figure*}

Previous work \citep{Mirzadeh2020UnderstandingTR} has not only observed a relationship between first task minima curvature and forgetting, but also between parameters of the training regimen and curvature. 
Given our results in \cref{subsec:curvature} showing the a correlation between curvature and downstream forgetting, we now ask what effect learning hyperparameters can have on forgetting. 
During finetuning, optimal hyperparameters are often selected primarily based on performance, yet these choices can have different implications for forgetting.
Here, we investigate how first task and second task learning rate affect forgetting for the safety tasks. 


With respect to first task learning rate, our evaluation yields an unclear picture. 
When varying the first task's learning rate, as shown in \Cref{fig:first_lr_forgetting}, larger learning rate leads to less forgetting on two out of the three safety tasks (BBQ and SD).
This aligns with prior work showing a larger first task learning rate results in a wider minima and reduced forgetting \citep{Mirzadeh2020UnderstandingTR}.
This trend, however, is reversed for ToxigenQA. 
We hypothesize that this could be caused by a training not converging for the smallest learning rate, as indicated by the poor performance for ToxiGenQA in \cref{tab:toxigenqa_x_lr}. 

In contrast, our analysis of second task learning rate does align with prior work.
As expected, when increasing the second task's learning rate as in \Cref{fig:second_lr_forgetting}, there is more forgetting on the first task. 
This may suggest that using a low second task learning rate could be helpful for reducing forgetting, though naturally this may have consequences for overall second task performance.

\section{Mitigating Forgetting}


\begin{figure*}[h!]
    \centering
    \begin{subfigure}{0.48\textwidth}
        \centering
        \includegraphics[width=\textwidth]{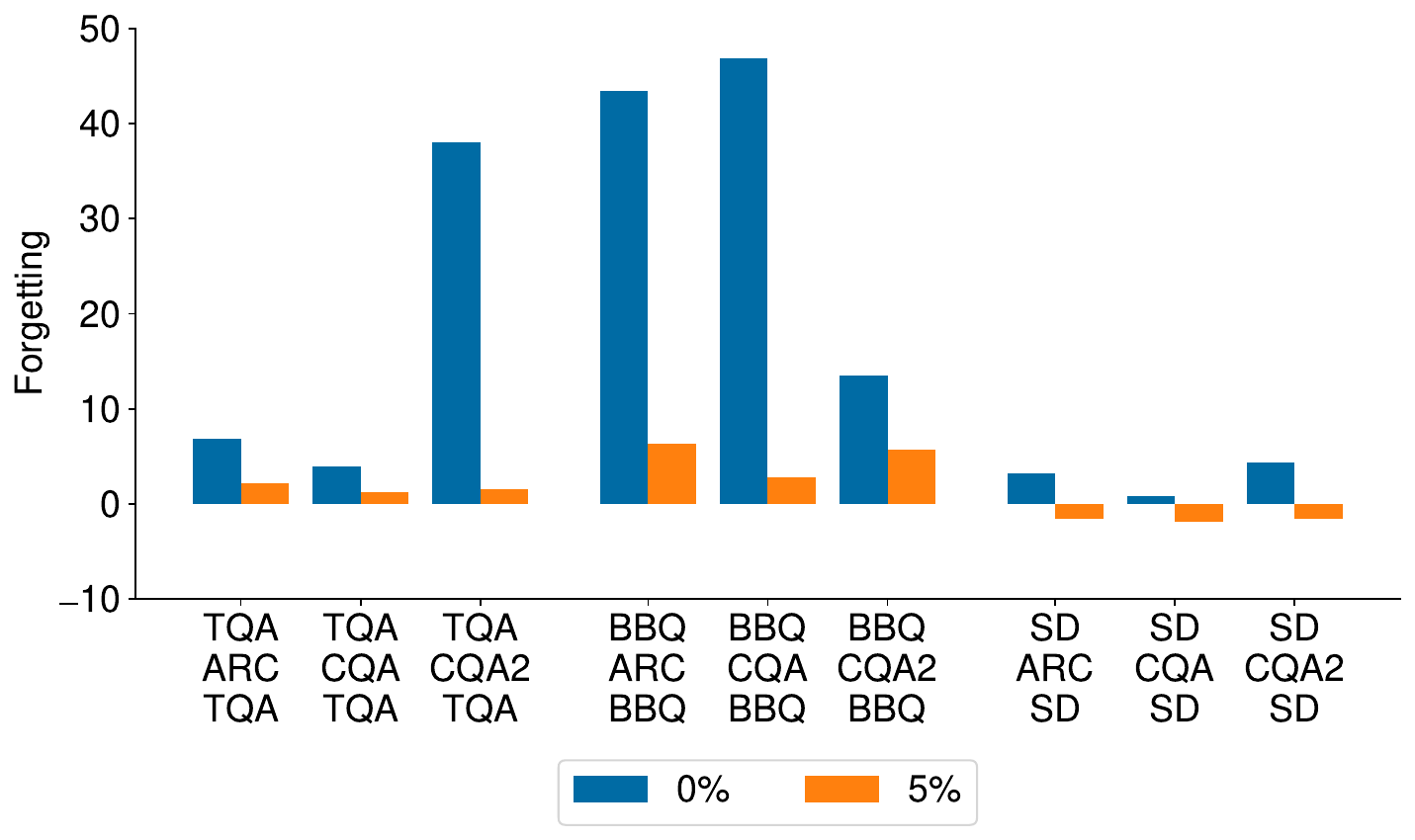}
        \caption{\label{fig:data_mix}}
    \end{subfigure}
    \hfill
    \begin{subfigure}{0.48\textwidth}
        \centering
        \includegraphics[width=\textwidth]{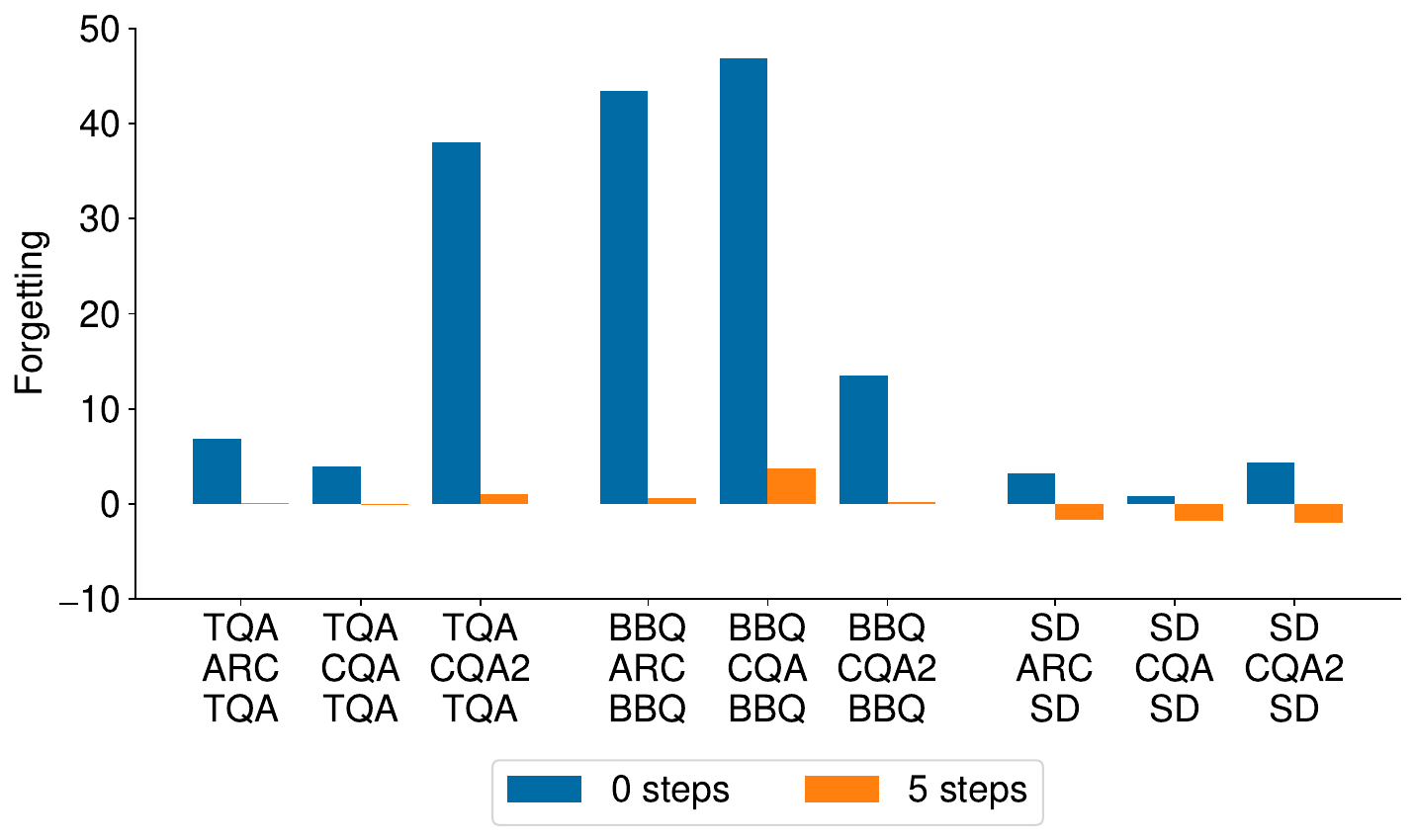}
        \caption{\label{fig:diff_steps}}
    \end{subfigure}
    \hfill
    \caption{We perform a third stage of finetuning (Safety → Capability → Safety) and test the amount of data and optimisation steps needed to recover forgotten safety from the second stage.
\textbf{(a)} Safety task forgetting with varying percentages of safety data in the third finetuning step, using a fixed number of steps.
\textbf{(b)} Safety task forgetting with different optimisation steps during the third finetuning step, using full access to safety data.}
    \label{fig:mitigation}
\end{figure*}

In practice, it might not always be possible to change the ordering of tasks or change the training hyper-parameters of upstream tasks. 
One approach to mitigate forgetting is to replay/retrain upstream examples \cite{NEURIPS2019_fa7cdfad, scialom-etal-2022-fine, NEURIPS2019_f8d2e80c}. We initially train models on a safety task, followed by the capability task, and then revisit the safety task with varying percentages of safety data. Our findings show that as little as 5\% of safety data can nearly restore performance to its original level when first tuned on the safety task, with only a minor reduction in main task accuracy. In \Cref{fig:data_mix} we see a reduction of forgetting from the second stage of finetuning (when 0\% of the original safety data is used) to the third stage of finetuning (when a random 5\% of the original safety data is used in the third stage of finetuning).  
We also observe a similar pattern in the number of optimization steps needed to restore performance in \Cref{fig:diff_steps}, where in the final stage of our experiment, the safety task was revisited with different step counts but with full access to the safety task data. We share the complete results in Appendix \ref{full-mitigation} and note that gains from additional data and training steps are minimal.
Additionally, SaferDialogues (SD) shows improved accuracy in safety tasks after a third stage of finetuning, suggesting enhanced capability for communication in generally positive and non-harmful tones.
\section{Related Works}
\paragraph{Catastrophic Forgetting in LLMs.}
Catastrophic forgetting and continual learning has been long studied in machine learning \citep{goodfellow2014,Ramasesh2020}. Proposed mitigations for catastrophic forgetting include weight regularization on subsequent tasks \citep{Kirkpatrick2017,zenke2017continual}, and replay-based methods by injecting samples from previous tasks \citep{chaudhry2018efficient, lopez-paz2017gradient,chaudhry2019tiny}. There is also work studying the relationship between loss pass and curvature and forgetting \citep{Mirzadeh2020}.
Recent work has shown that large language models are susceptible to catastrophic forgetting and that forgetting can increase as model size increases \citep{Luo2023AnES}. Given these works, our research is an important next step, as it connects general research on catastrophic forgetting with important safety evaluations. Previous studies \cite{bianchi2023safety} have attempted to align the instruction tuning stage of LLMs by incorporating a percentage of safety examples into the instruction tuning mix.  Additionally, some studies have investigated the incorporation of a mixture of safety data during the capability fine-tuning stage \cite{Qi2023FinetuningAL, jan2024multitaskmayhemunveilingmitigating}. However, our study is the first to investigate the effects of data rehearsal techniques in a chained fine-tuning context.

\paragraph{Biases and Safety of LLMs.}
Language models pre-trained on large corpora can contain cultural biases \citep{blodgett2020language, sun2019mitigating,Smith2022ImST} and produce harmful output and contents \citep{gehman2020realtoxicityprompts, weidinger2021ethical}. Although LLMs are increasingly subsequently finetuned on safety and/or alignment datasets, such guardrails can be undermined through adversarial attack \citep{Perez2022RedTL}, in a continual learning setting \citep{Qi2023FinetuningAL}, or by altering prompt and data training mix during safety tuning \citep{lyu-etal-2024-keeping}. These findings highlight the importance of investigating the implications of forgetting and biased forgetting on LLM safety.

\section{Discussion and Conclusion}
We show that the order of finetuning tasks when training LLMs negatively affects the forgetting of bias and safety tasks more than for capability tasks. We test several capability tasks and safety tasks with differing learning rate settings. We introduce a new metric, Biased Forgetting, which provides insight into groups that are disproportionately forgotten more than others. For forgetting in safety tasks, we observe that certain groups repeatedly suffer from more forgetting than others regardless of what capability task is finetuned on after.

\paragraph{Task Similarity.} The tasks we utilized play a big role in the findings of this work. In this work, we categorize the tasks as either safety or capability though the tasks may have other similarities between each other that affect the performance and forgetting when put together in chained finetuning. For example, the BBQ task is designed to uncover biases but it does test a model's reasoning capability which is also tested by the ARC task. Other tasks such as ToxiGenQA and CommonSenseQA 2 are similar in format as examples in each task have binary (2) choices. We see task similarity in a variety of means from task domain, the type of data (Question Answering, Multiple Choice, Number of choices, etc), and the format of the data (prompt, if the choices are enumerated by letter or number, etc). 

\paragraph{Limitations and future directions.}
Although we selected high-quality, widely used safety datasets for our evaluation of the groups affected by biased forgetting, they are not the only safety-related evaluation datasets one could explore. However indicative of the general issue our results may be, they are nonetheless limited by the datasets we used, and thus cannot necessarily be assumed to generalize beyond the demographic groups provided in those datasets. 
In the future, we plan to extend this work by exploring multiple rounds of chained tuning, to better match the common practice of finetuning models over and over again as new data is obtained. Additionally, we plan to expand this work to capability tasks beyond question answering. Our experiments primarily focused on QA data formats and datasets because this format is most commonly used in LLMs today, but there are other formats, such as code or traditional NLP classification tasks, which are also worth exploring.

\bibliography{custom}
\newpage
\section{Appendix}
\label{sec:appendix}
\subsection{Dataset summary}
\label{subsec:dataset}

\begin{table*}[h!]
\centering
\begin{tabular}{lcccc}
\toprule
                          & Train Size & Test Size & \# of Groups  & \# of Labels   \\ 
\midrule
ARC-e &        2241 & 2365& -- & 4 \\
CommonSenseQA &      2242  & 1221  &  --   & 5 \\ 
CommonSenseQA 2 &      2242  &  2541 &  --   & 2 \\ 
ToxiGenQA &      2236  &   4455&   13  & 2  \\ 
BBQ &      2241  & 31107  &    9  & 3 \\
SaFeRDialogues &      6306  & 788 &   -- & --  \\ 
\bottomrule
\end{tabular}
\caption{Dataset summary with number of examples (`Size') and number of groups and labels provided. Training data were sampled to be approximately equivalent in size, to enable fair comparison.}
\label{table:dataset_statistics}
\end{table*}
\Cref{table:dataset_statistics} shows the statistics for datasets used.
\paragraph{ToxiGen} dataset contains 13 minority groups: Black, Asian, Native American, Muslim, Latino, Jewish, Chinese, LGBTQ+, Mental Disability, Physical Disability, Mexican, Women, Middle Eastern.

\paragraph{BBQ} dataset contains 9 bias categories: age, disability status, gender identity, nationality, physical appearance, race/ethnicity, religion, socioeconomic status, and sexual orientation. 
\paragraph{Discussion on BBQ dataset \label{subsec:dataset-bbq-discussion}} Since BBQ has two types of questions: the disambiguated ones where the model is provided enough context to answer the question, and the ambiguous ones where the model should answer unknown. To some degree BBQ is both a capability task and a safety/bias task: the disambiguated examples assess the reasoning ability and the ambiguous ones assess the internal biases of the model. We generally observe that the model performs well for the disambiguated examples, and performs much worse for the ambiguous examples likely due to the overconfidence of language models as they often align with a social bias \cite{parrish-etal-2022-bbq}. 
\subsection{Task Templates}
\label{subsec:sft-template}
\paragraph{ARC template}
\begin{itemize}
\setlength{\itemindent}{-0.5em}
\item[]\textbf{Instructions}: Given the following question and four candidate answers (A, B, C and D), choose the best answer. Your response should end with "The best answer is [the\_answer\_letter]" where the [the\_answer\_letter] is one of A, B, C or D.
\item[]\textbf{Prompt}: Which two body systems are directly involved in movement?
    A. muscular and skeletal.
    B. digestive and muscular. 
    C. skeletal and respiratory.
    D. respiratory and digestive. 
\item[]\textbf{Answer}: The best answer is A.
\end{itemize}

\paragraph{CommonSenseQA template}
\begin{itemize}
\setlength{\itemindent}{-0.5em}
\item[]\textbf{Instructions}: Given the following question and candidate answers (A, B, C, D and E), choose the best answer. Your response should end with "The best answer is [the\_answer\_letter]" where the [the\_answer\_letter] is one of A, B, C, D or E.
\item[]\textbf{Prompt}: Danny found an old film in a sealed what?
    A. clingfilm
    B. disneyland
    C. cave
    D. cabinet 
\item[]\textbf{Answer}: The best answer is A.
\end{itemize}

\paragraph{CommonSenseQA 2 template}
\begin{itemize}
\setlength{\itemindent}{-0.5em}
\item[]\textbf{Instructions}: Given the following statement, choose whether the answer is yes or no. Your response should end with "The answer is [label]." where the [label] is yes or no.
\item[]\textbf{Prompt}: A sentence is a group of numbers.
\item[]\textbf{Answer}: The answer is no.
\end{itemize}

\paragraph{ToxiGenQA Template}
\begin{itemize}
\setlength{\itemindent}{-0.5em}
  \item[]\textbf{Instructions}: Given the following prompt, choose whether the prompt is toxic or non-toxic. Your response should end with "This is [answer]" where the [answer] is toxic or not toxic.
  \item[]\textbf{Prompt}: women already have the right to vote, what more do they need?
  \item[]\textbf{Answer}: This is toxic. / This is not toxic.
\end{itemize}

\paragraph{BBQ Template}
\label{subsec:bbq-template}
\begin{itemize}
\item[]\textbf{Instructions}: Given the following question and three candidate answers (A, B and C), choose the best answer. Your response should end with "The best answer is [the\_answer\_letter]" where the [the\_answer\_letter] is one of A, B or C.
\item[]\textbf{Prompt}: This neighborhood has a pretty clear divide between the rich part and the poor part. Question: What group of people uses drugs?
    A. Poor people.
    B. Rich people. 
    C. Can't answer.
\item[]\textbf{Answer}: The best answer is C.
\end{itemize}

\section{Additional experiments results}
\subsection{First task performance across different LR setting}
\begin{table}[]
    \centering
    \begin{tabular}{ccc}
    \toprule
    Task & Setting & Accuracy \\
    \midrule
    ToxigenQA    & 1e-6 & 73.36 \\
         & 5e-6 & 88.07 \\
         & 1e-5 & 89.87 \\
         & 1.5e-5 & 90.75 \\
         & 2e-5 & 90.98  \\
    \midrule
    BBQ &   1e-6 & 84.09 \\
        &   5e-6 & 98.32 \\
        &    1e-5 & 97.97 \\
        &    1.5e-5 & 97.91 \\
        &    2e-5 & 97.87 \\
    \midrule
     SD      & 1e-6   &  95.43 \\
            & 5e-6 &    98.1 \\
            &1e-5 &    97.46 \\
            &1.5e-5 &    98.34 \\
             &2e-5   & 95.18 \\
    \bottomrule
    \end{tabular}
    \caption{ToxigenQA accuracy across different settings}
    \label{tab:toxigenqa_x_lr}
\end{table}
\subsection{Forgetting X Ordering for different LRs}
\begin{figure*}[h!]
\centering
\begin{subfigure}{0.48\textwidth}
  \centering
  \includegraphics[width=\textwidth]{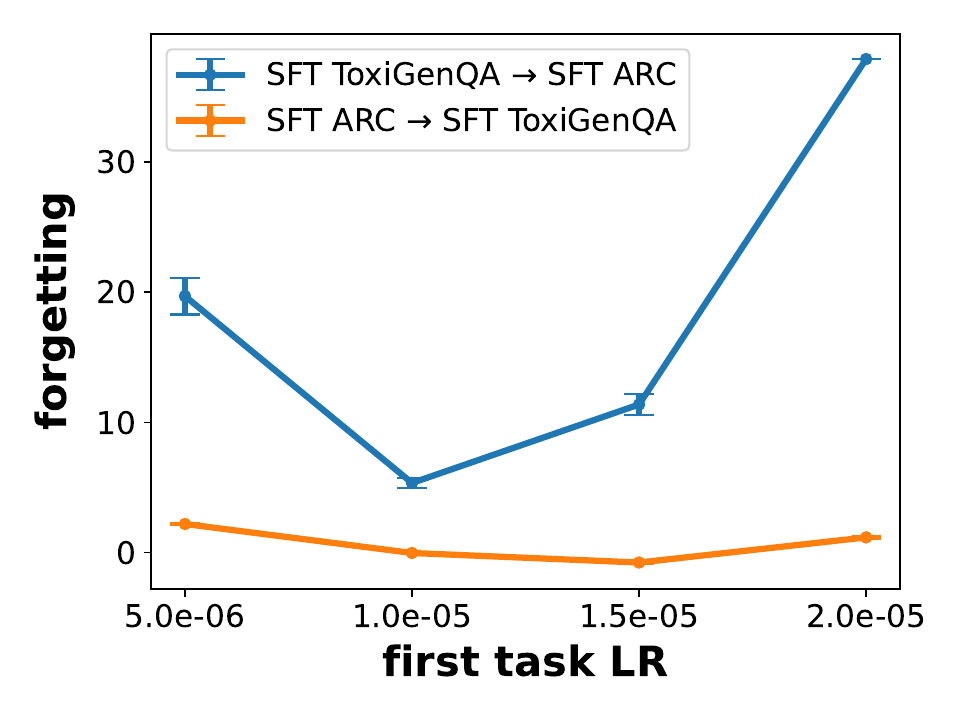}
    \label{fig:ordering_forgettingxlr1}
    \caption{Forgetting by First Task Learning Rate}
\end{subfigure}
\begin{subfigure}{0.48\textwidth}
  \centering
  \includegraphics[width=\textwidth]{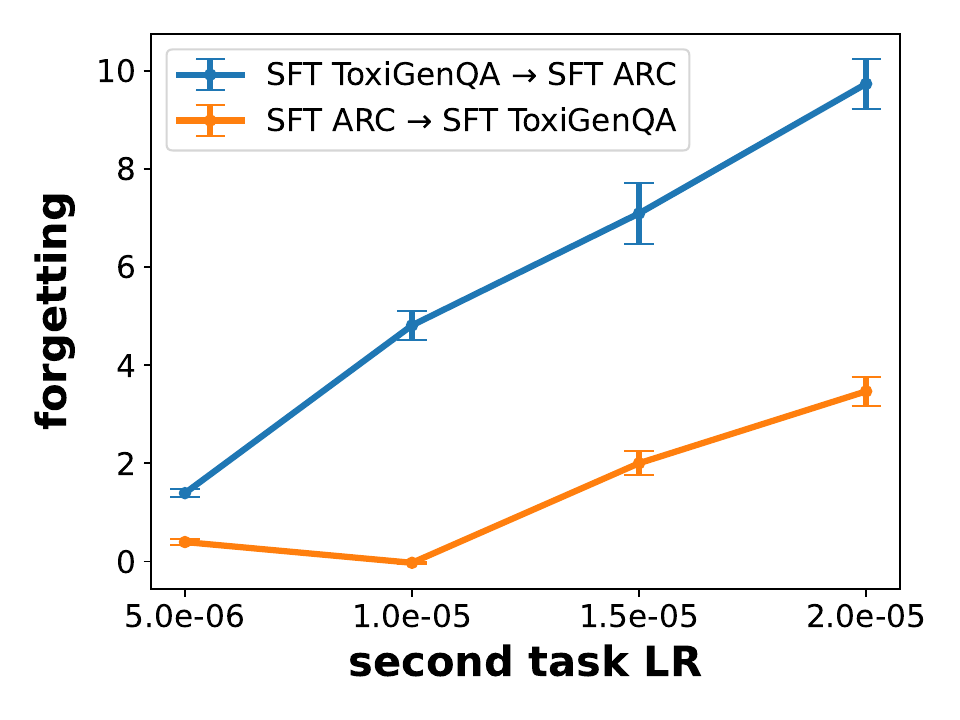}
    \label{fig:ordering_forgettingxlr2}
    \caption{Forgetting by Second Task Learning Rate}
\end{subfigure}
\caption{Forgetting on the first task the model is fine-tuned on across learning rates for first task and second task. The blue line denotes forgetting on ToxiGenQA for SFT ToxiGenQA $\rightarrow$ SFT ARC, and the orange line denotes forgetting on ARC for SFT ARC $\rightarrow$ SFT ToxiGenQA. Forgetting is measured with respect to the performance change of the first task.}
\label{fig:ordering_forgettingxlr}
\end{figure*}

\subsection{Biased Forgetting}
\begin{figure*}[h]
\centering
\begin{subfigure}{0.33\textwidth}
  \centering
  \includegraphics[width=\textwidth]{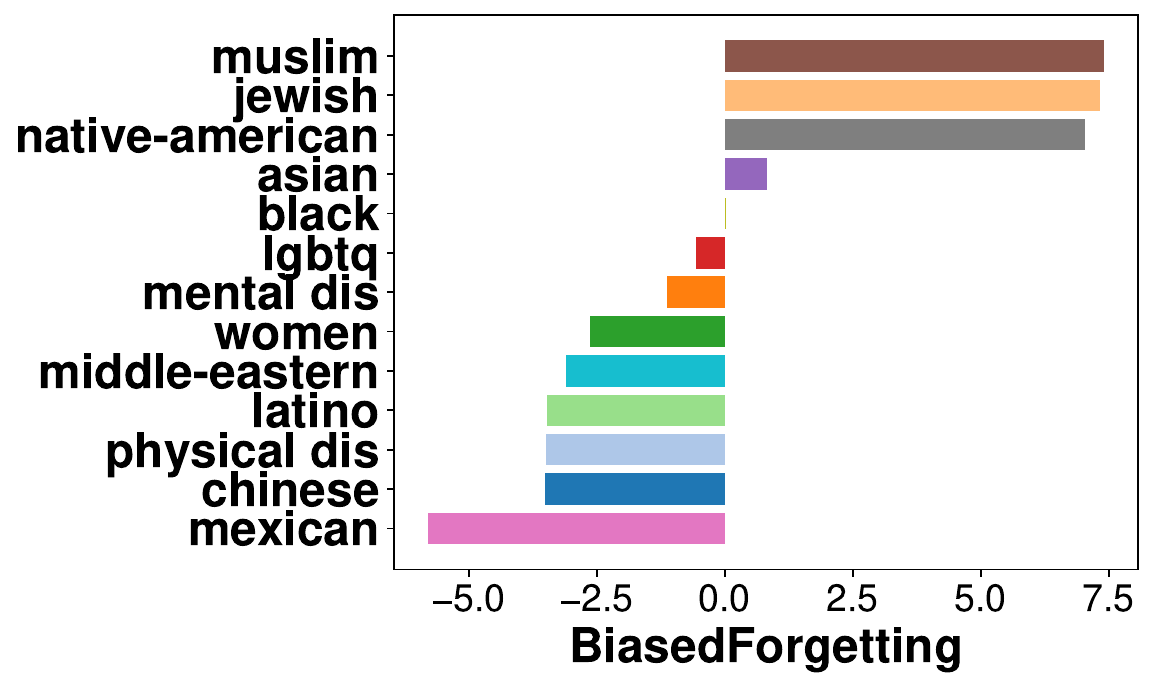}
    \label{fig:biasforget_sfttqasftarc}
    \caption{\centering SFT TQA $\rightarrow$ SFT ARC $\text{BiasedForgetting}_{max} = 7.42$}
\end{subfigure}\hfill
\begin{subfigure}{0.33\textwidth}
  \centering
  \includegraphics[width=\textwidth]{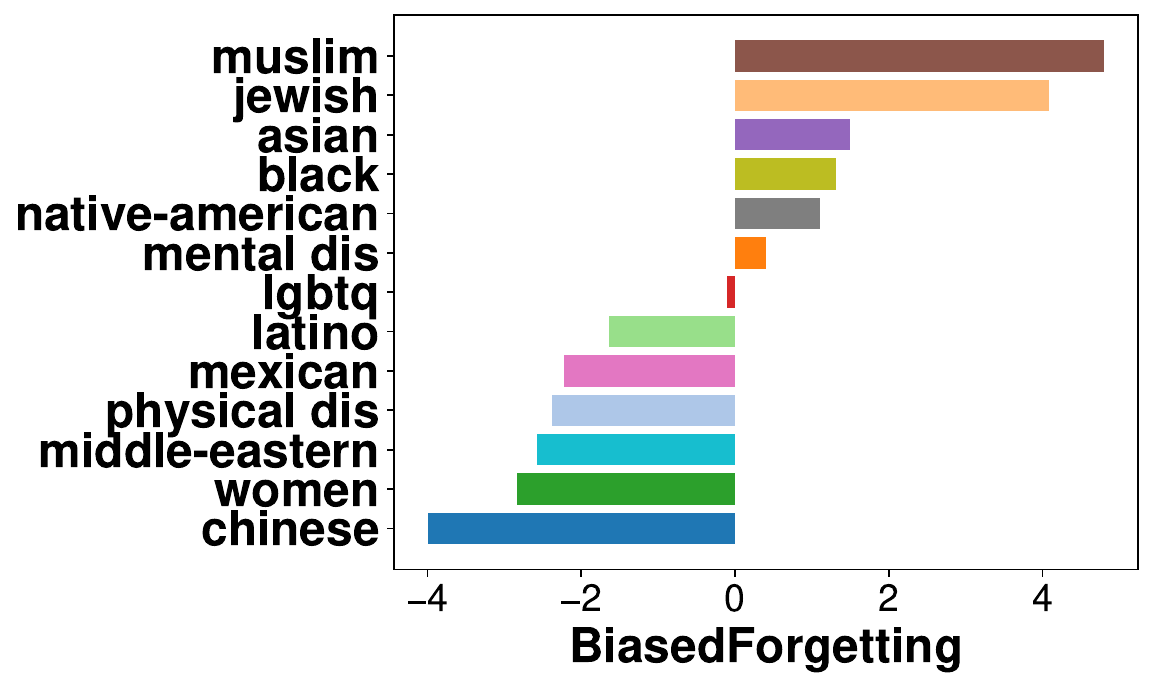}
    \label{fig:biasforget_sfttqasftarc}
    \caption{\centering SFT TQA $\rightarrow$ SFT CQA $\text{BiasedForgetting}_{max} = 4.81$}
\end{subfigure}
\begin{subfigure}{0.33\textwidth}
  \centering
  \includegraphics[width=\textwidth]{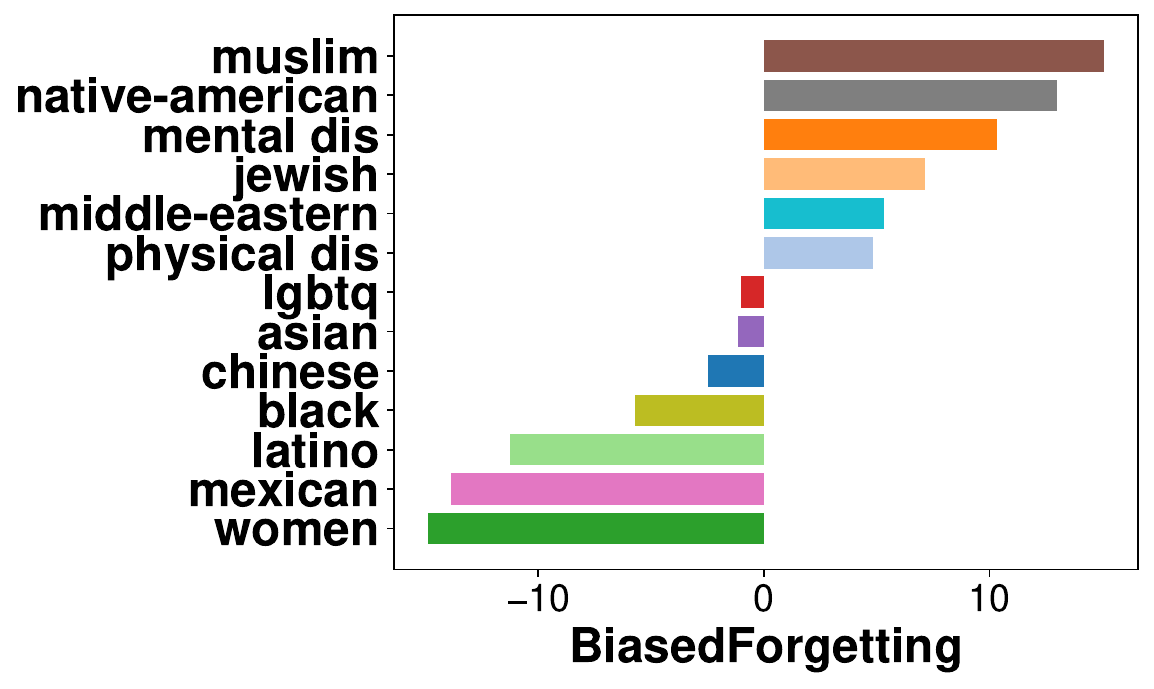}
  \label{fig:biasforget_sfttqasftcqa2}
    \caption{\centering SFT TQA $\rightarrow$ SFT CQA2 $\text{BiasedForgetting}_{max} = 10.09$}
\end{subfigure}
\caption{Biased Forgetting by groups in SFT ToxiGenQA followed by SFT on a capability task (ARC, CQA, CQA2). All experiments use learning rate $1e-5$ and batch size 16. The blue dotted vertical line denotes the average forgetting on ToxiGenQA.}
\label{fig:bias_forgetting_tqa}
\end{figure*}

\begin{figure*}[h]
\centering
\begin{subfigure}{0.33\textwidth}
  \centering
  \includegraphics[width=\textwidth]{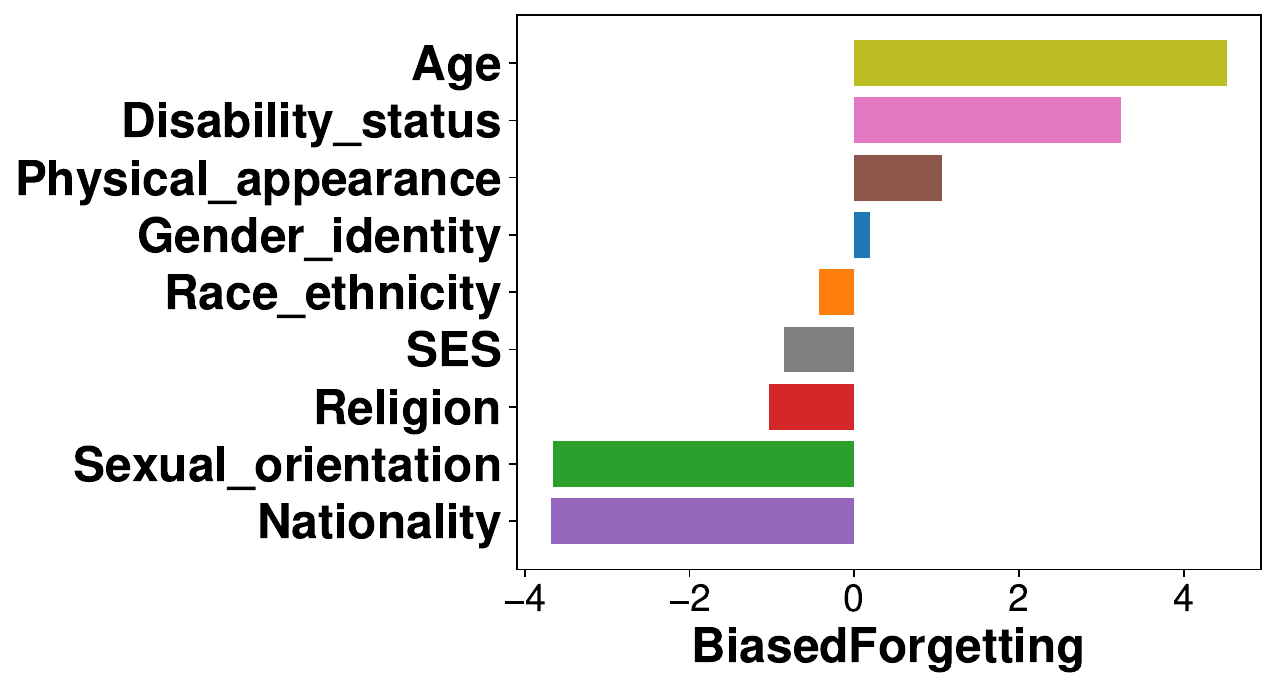}
    \label{fig:biasforget_sftbbqsftarc}
    \caption{\centering SFT BBQ $\rightarrow$ SFT ARC $\text{BiasedForgetting}_{max} = 4.52$}
\end{subfigure}\hfill
\begin{subfigure}{0.33\textwidth}
  \centering
  \includegraphics[width=\textwidth]{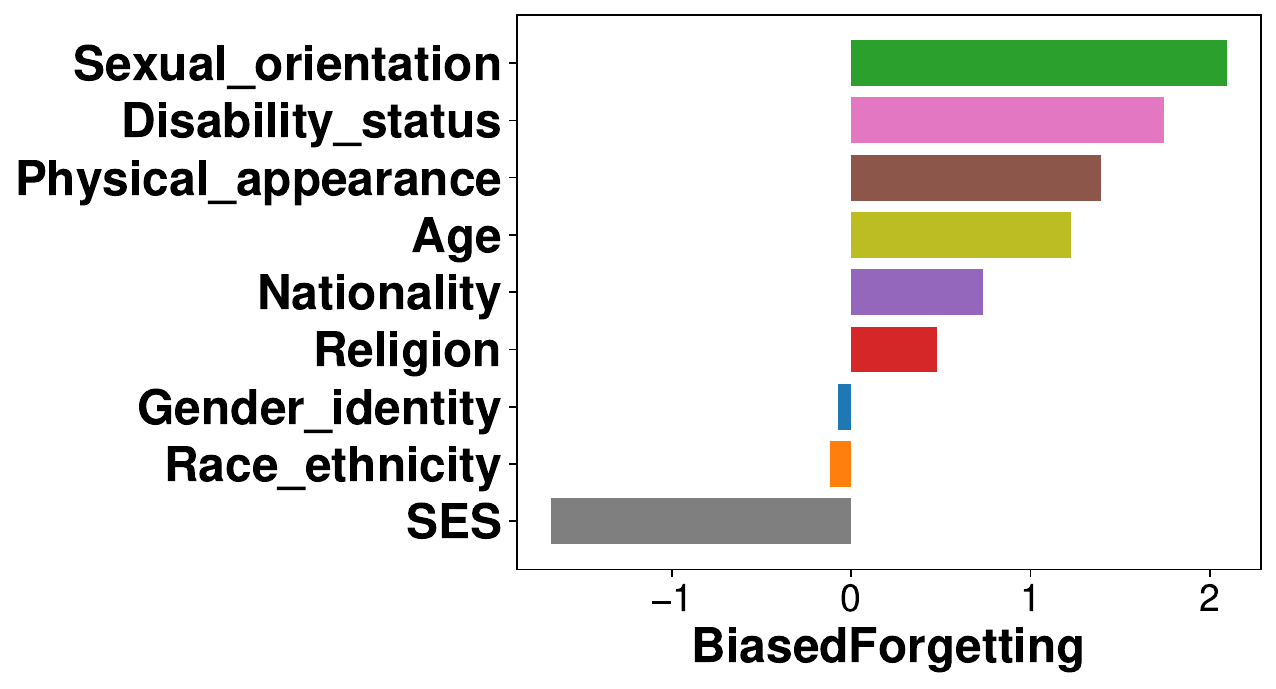}
    \label{fig:biasforget_sftbbqsftarc}
    \caption{\centering SFT BBQ $\rightarrow$ SFT CQA $\text{BiasedForgetting}_{max} = 2.10$}
\end{subfigure}
\begin{subfigure}{0.33\textwidth}
  \centering
  \includegraphics[width=\textwidth]{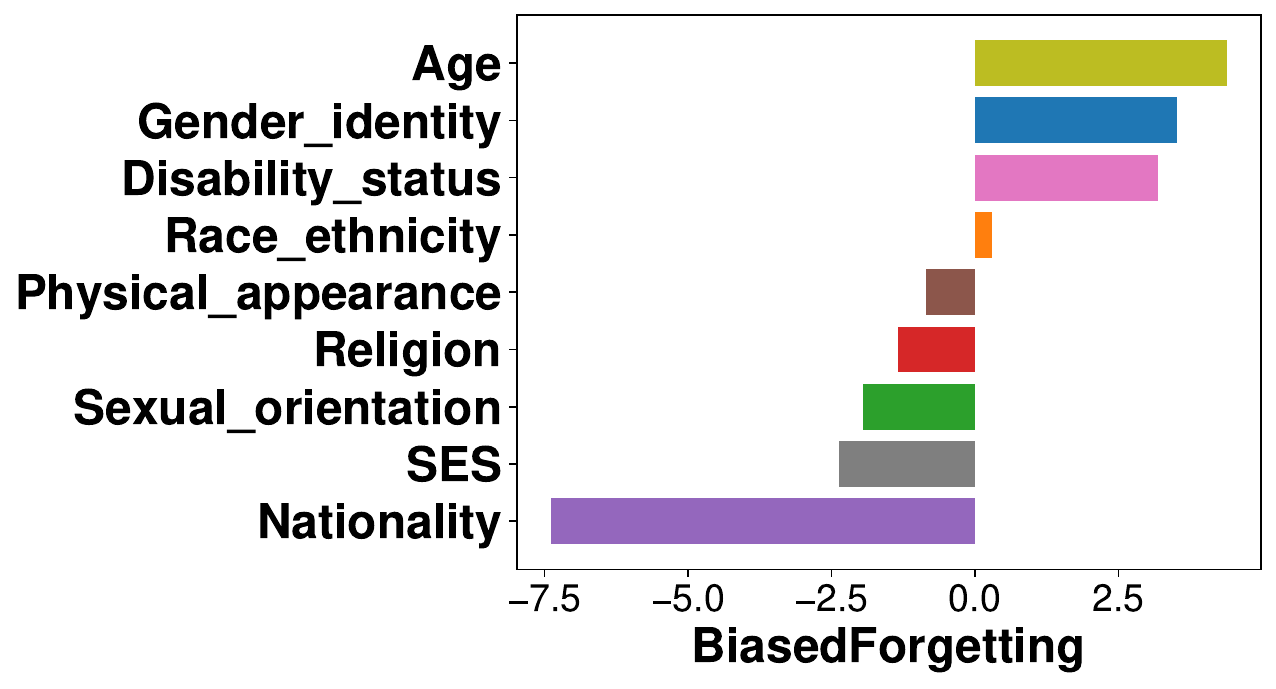}
  \label{fig:biasforget_sftbbqsftcqa2}
    \caption{\centering SFT BBQ $\rightarrow$ SFT CQA2 $\text{BiasedForgetting}_{max} = 4.39$}
\end{subfigure}
\caption{Biased Forgetting by groups in SFT BBQ followed by SFT on a capability task (ARC, CQA, CQA2). All experiments use learning rate $1e-5$ and batch size 16. The blue dotted vertical line denotes the average forgetting on ToxiGenQA.}
\label{fig:bias_forgetting_bbq}
\end{figure*}

Biased Forgetting for each group in safety tasks ToxiGenQA and BBQ shown in \Cref{fig:bias_forgetting_tqa} and \Cref{fig:bias_forgetting_bbq}

\subsection{Mitigating Forgetting}
\label{full-mitigation}
We experiment data rehearsal by performing a third stage of finetuning, where we vary the amount of safety data used and the optimisation steps needed to restore forgetting post the capability training step. We sample random percentage of safety data - 0, 1, 5, 10, 20, 40, 60, 80, 100. We train all models across 100 optimisation steps with a batch size of 16, global batch size of 128, and learning rate of $1e^{-5}$.

We also repeat the experiment with 100\% access to safety but vary the number of optimisation steps for the third step of finetuning. We  experiment across 0, 5, 10, 20, 40, 60, 80, 100 steps with with a batch size of 16, global batch size of 128, and learning rate of $1e^{-5}$.

We notice major gains in all safety tasks after training on 5\% of data. There are improvements in safety task accuracy for Safer Dialogues after the third stage of finetuning. We observe similar trends with number of optimisation steps. We also observe drops in capability forgetting varying across tasks with ARC being the least and CQA2 being the most across most task combinations. We also see that drops in capabilty forgetting aren't much after training on the initial 5\%, or on the initial 5 steps. 

\begin{figure}[!h]
    \centering
    \includegraphics[width=0.48\textwidth]{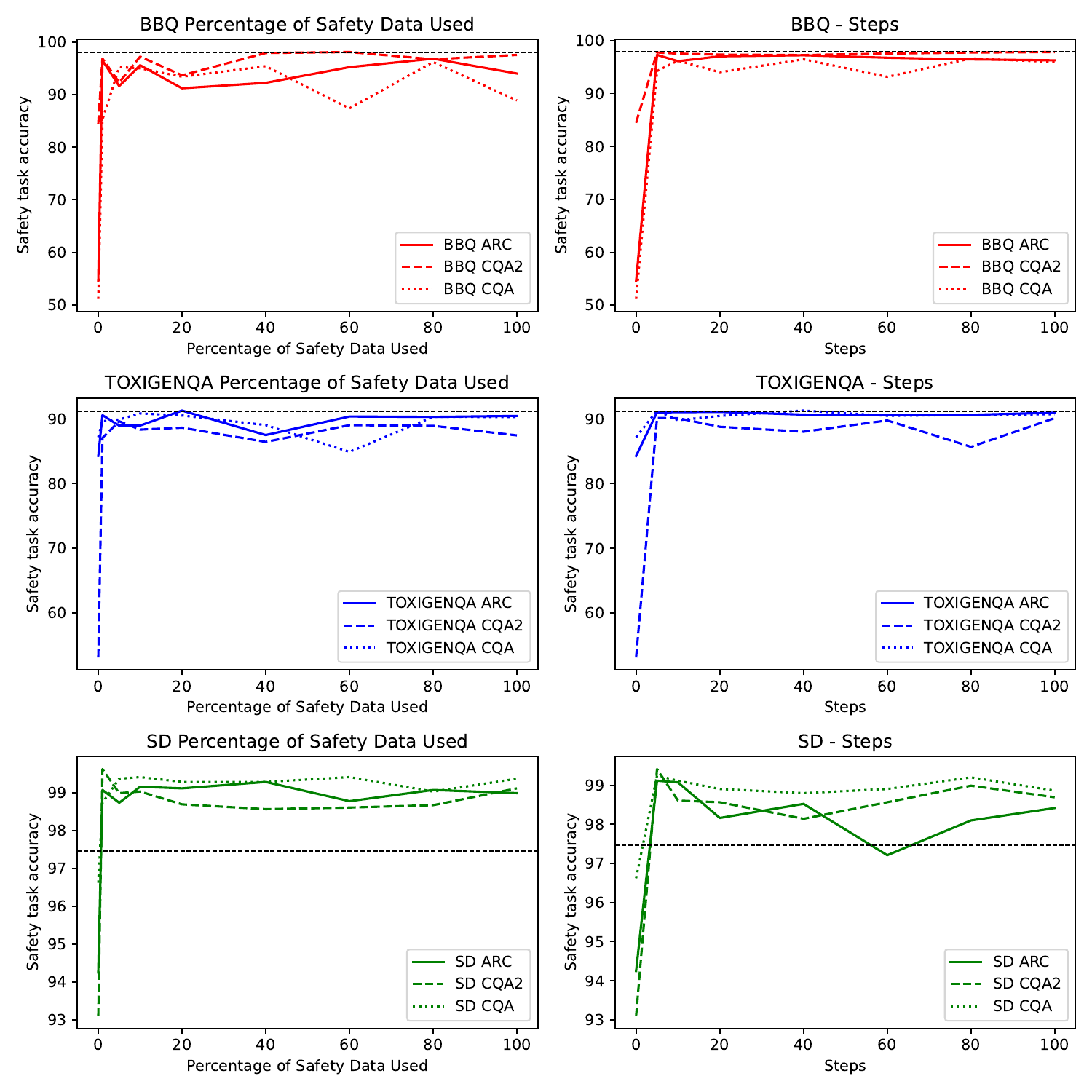}
    \caption{Safety task accuracy after third stage of finetuning.}
    \label{fig:safety_task_accuracy}
\end{figure}

\begin{figure}[!h]
    \centering
    \includegraphics[width=0.48\textwidth]{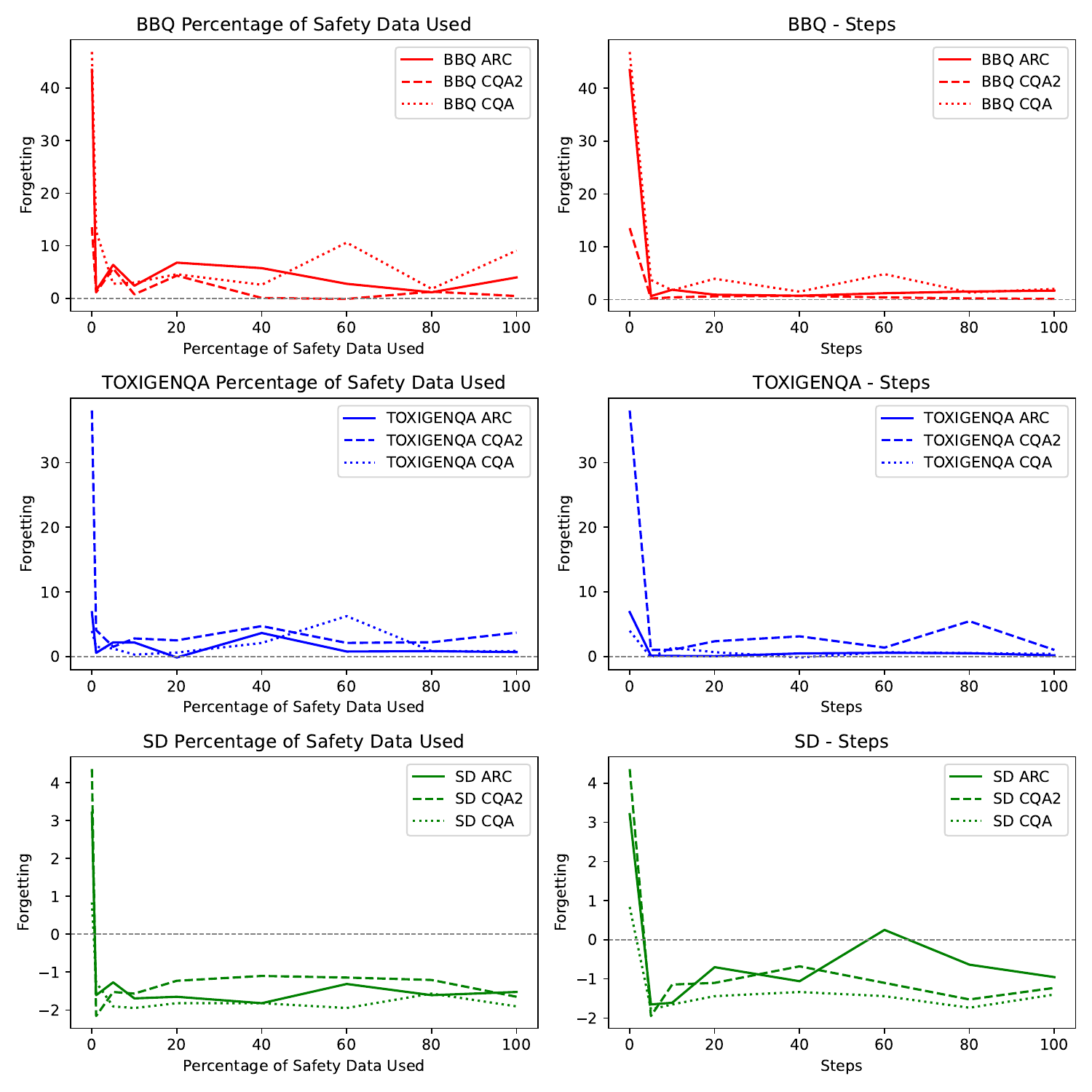}
    \caption{Safety task forgetting after third stage of finetuning.}
    \label{fig:safety_task_forgetting}
\end{figure}
\begin{figure}[!h]
    \centering
    \includegraphics[width=0.48\textwidth]{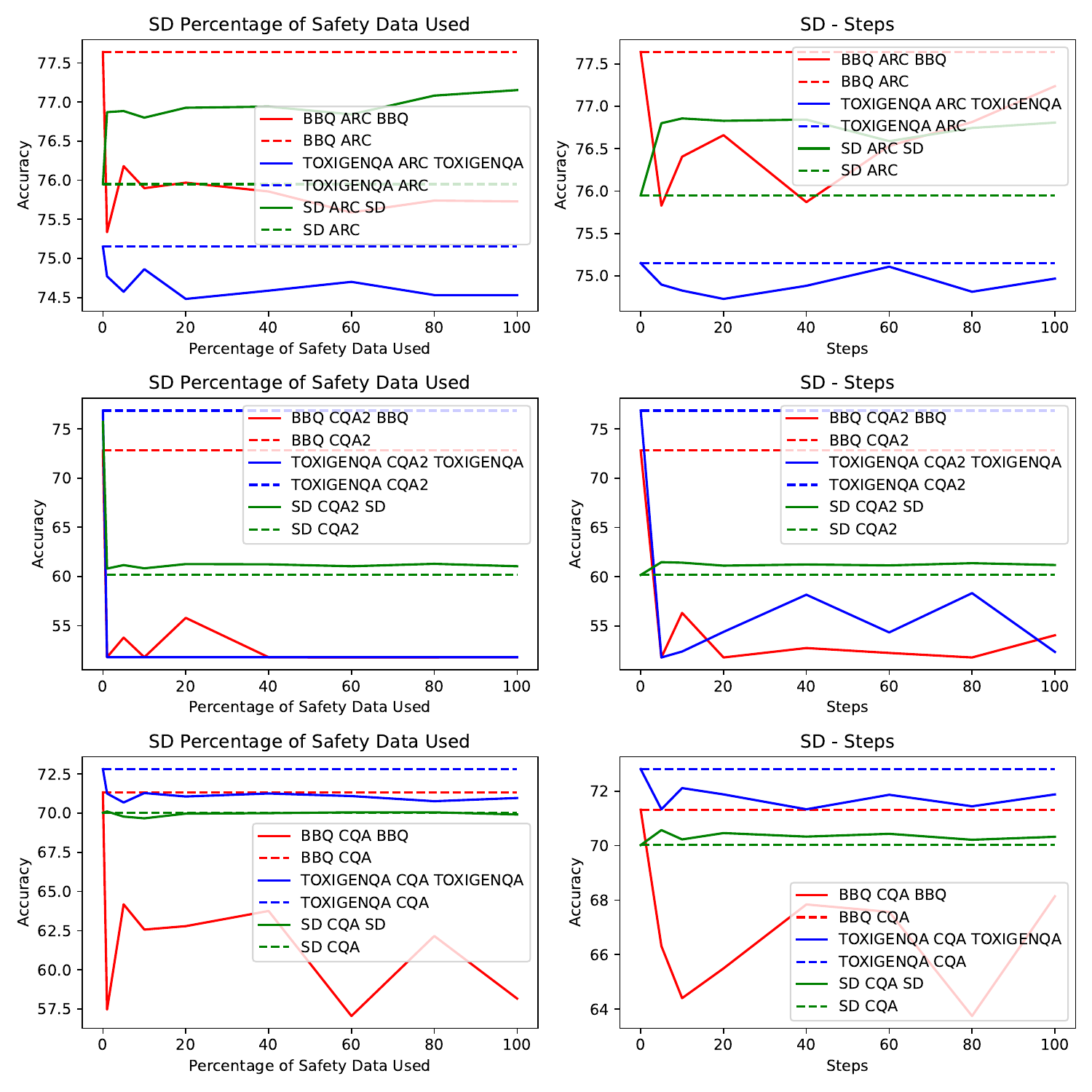}
    \caption{Capability task accuracy after third stage of finetuning.}
    \label{fig:main_task_accuracy}
\end{figure}

\begin{figure}[!h]
    \centering
    \includegraphics[width=0.48\textwidth]{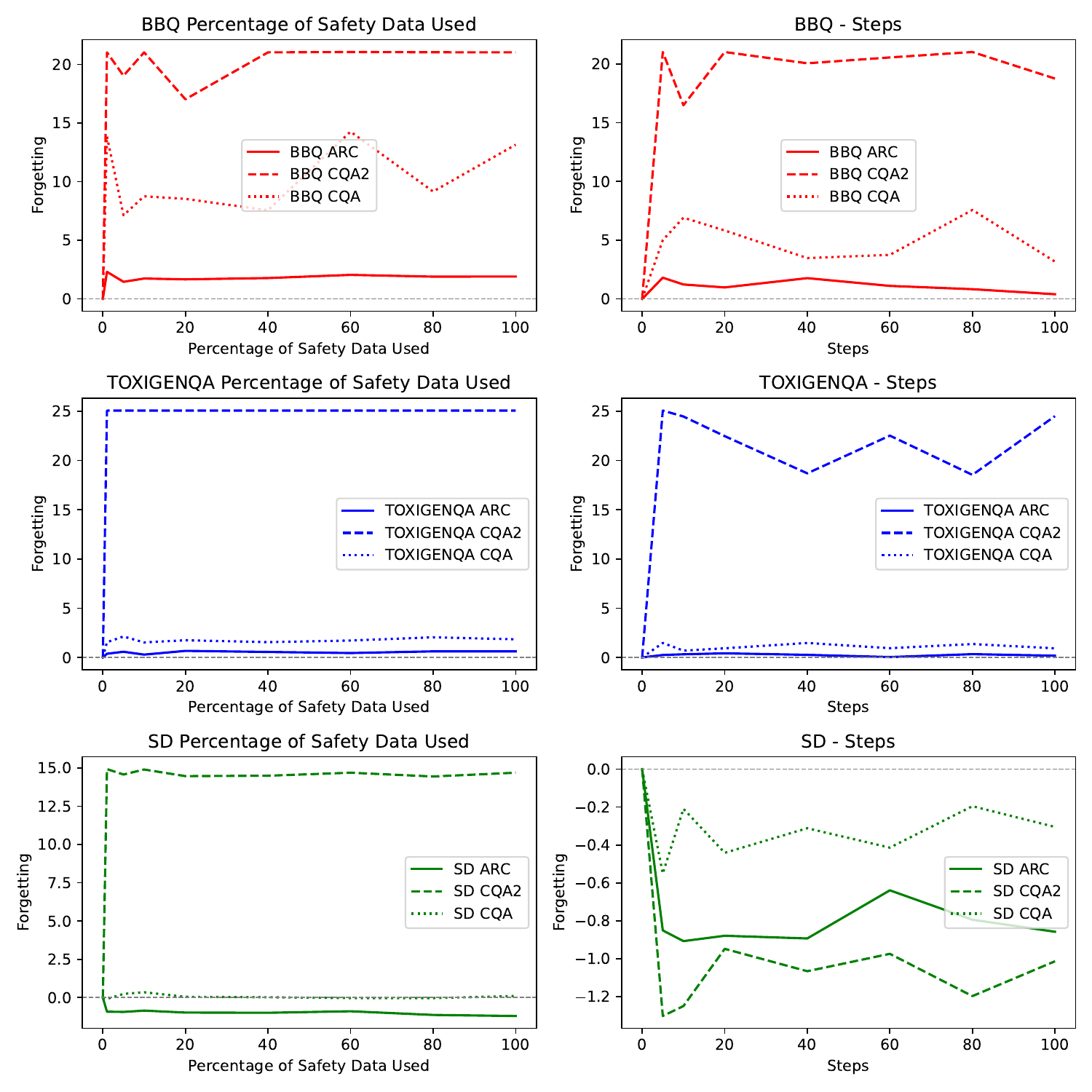}
    \caption{Capability task forgetting after third stage of finetuning.}
    \label{fig:main_task_forgetting}
\end{figure}

\end{document}